\documentclass[10pt,twocolumn,letterpaper]{article}

\usepackage{cvpr}              

\usepackage{graphicx}
\usepackage{amsmath}
\usepackage{amssymb}
\usepackage{booktabs}
\usepackage{xcolor}
\usepackage{paralist}
\usepackage{tabularx}
\definecolor{LightGray}{gray}{0.98}
\usepackage{ulem}

\makeatletter
\@namedef{ver@everyshi.sty}{}
\makeatother
\usepackage{pgfplots} 
\pgfplotsset{compat=1.16}

\usepackage[pagebackref,breaklinks,colorlinks]{hyperref}

\usepackage[capitalize]{cleveref}
\crefname{section}{Sec.}{Secs.}
\Crefname{section}{Section}{Sections}
\Crefname{table}{Table}{Tables}
\crefname{table}{Tab.}{Tabs.}

\newcommand{\eb}{EditBench\xspace}
\newcommand{\imrm}{IM$_{\text{RM}}$\xspace}
\newcommand{\im}{IM\xspace}
\newcommand{\imagenator}{Imagen Editor\xspace}
\newcommand{\imagenatorrm}{Imagen Editor$_{\text{RM}}$\xspace}
\newcommand{\dalle}{DALL-E 2\xspace}
\newcommand{\dl}{DL2\xspace}
\newcommand{\stablediffusion}{Stable Diffusion\xspace}
\newcommand{\sd}{SD\xspace}

\newcommand{\task}{text-guided image inpainting\xspace}
\newcommand{\tasklong}{text-guided image inpainting\xspace}
\newcommand{\tasklongcapitals}{Text-Guided Image Inpainting\xspace}

\newcommand{\full}{\emph{Full}\xspace}
\newcommand{\simple}{\emph{Mask-Simple}\xspace}
\newcommand{\rich}{\emph{Mask-Rich}\xspace}

\newcommand{\bfit}[1]{\textbf{\textit{#1}}}

\newcommand{\vb}{$\mid$}

\newcommand{\aos}[3]{`a=\emph{#1}\vb o=\emph{#2}\vb s=\emph{#3}'}

\newlength\replength
\newcommand\repfrac{.66}

\newcommand\rulewidth{.8pt}
\newcommand\tdashfill[1][\repfrac]{\cleaders\hbox to \replength{%
  \smash{\rule[\arraystretch\ht\strutbox]{\repfrac\replength}{\rulewidth}}}\hfill}

\def\cvprPaperID{12107} 
\def\confName{CVPR}
\def\confYear{2023}

\begin{document}

\normalem


\title{\vspace{-6mm}\imagenator{} and \eb{}: Advancing and Evaluating\\\tasklongcapitals\vspace{-3mm}}

\author{
{\normalsize \textbf{Su Wang$^*$ \quad Chitwan Saharia$^*$ \quad Ceslee Montgomery$^*$}}\\
{\normalsize \textbf{Jordi Pont-Tuset \quad Shai Noy \quad Stefano Pellegrini \quad Yasumasa Onoe}}\\
{\normalsize \textbf{Sarah Laszlo \quad David J. Fleet \quad Radu Soricut \quad Jason Baldridge}}\\
{\normalsize \textbf{Mohammad Norouzi$^\dagger$ \quad Peter Anderson$^\dagger$ \quad William Chan$^\dagger$}\vspace{5px}}\\Google Research}


\twocolumn[{%
\renewcommand\twocolumn[1][]{#1}%
\maketitle
\begin{center}
    \centering
    \captionsetup{type=figure}
    \vspace{-3mm}      
    \includegraphics[width=0.95\linewidth]{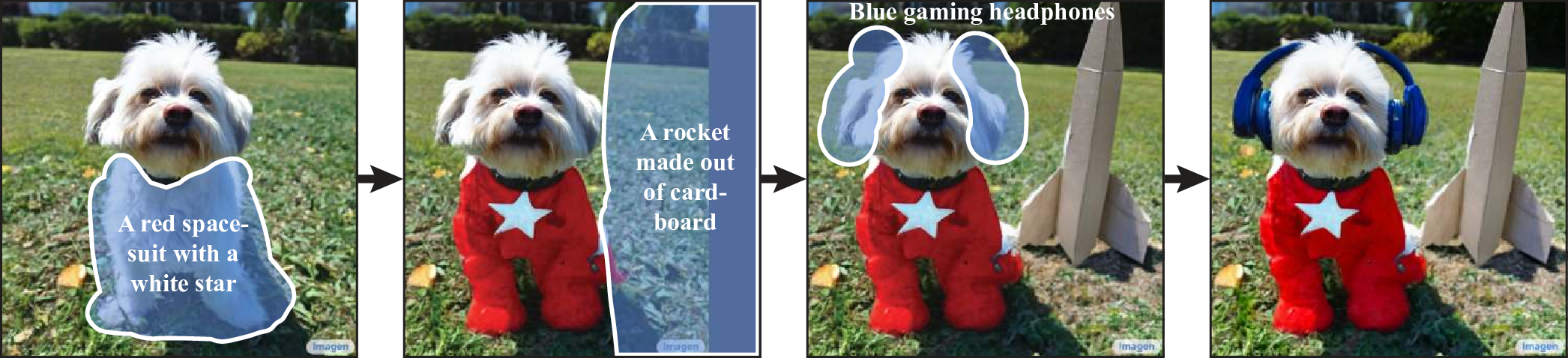}
    \captionof{figure}{\textbf{A sequence of edits by \imagenator}. Given an \emph{image}, a \emph{user defined mask}, and a \emph{text prompt}, \imagenator makes localized edits to the designated areas. The model meaningfully incorporates the user's intent and performs photorealistic edits.
    }
    \vspace{1mm}
    \label{fig:concept}
\end{center}%
}]

\maketitle


\begin{abstract}{\let\thefootnote\relax\footnote{$^*$Equal contribution. $^\dagger$Equal advisory contribution.}}

\noindent
Text-guided image editing can have a transformative impact in supporting creative applications.
A key challenge is to generate edits that are faithful to input text prompts, while consistent with input images.
We present \textbf{\imagenator{}}, a cascaded diffusion model built, by fine-tuning Imagen~\cite{sahariac-imagen} on \tasklong. 
\imagenator's edits are faithful to the text prompts, which is accomplished by using object detectors to propose \textit{inpainting masks} during training. 
In addition, \imagenator{} captures fine details in the input image by conditioning the cascaded pipeline on the original high resolution image.
To improve qualitative and quantitative evaluation, we introduce \textbf{\eb{}}, a systematic benchmark for \tasklong.
\eb{} evaluates inpainting edits on natural and generated images exploring objects,
attributes,
and scenes.
Through extensive human evaluation on \eb{}, we find that object-masking during training leads to across-the-board improvements in text-image alignment -- such that \imagenator{} is preferred over \dalle{}~\cite{ramesh-dalle2} and \stablediffusion{}~\cite{rombach-cvpr-2022} -- and, as a cohort, these models are better at object-rendering than text-rendering, and handle material/color/size attributes better than count/shape attributes. 




\vspace{-2mm}


\end{abstract}


\section{Introduction}




\noindent
Text-to-image generation has seen a surge of recent interest~\cite{ramesh-dalle2, sahariac-imagen, yu-parti, rombach-cvpr-2022, ernie-vilg}.
While these generative models are surprisingly effective, users with specific artistic and design needs 
do not typically obtain the desired outcome in a single interaction with the model.
Text-guided image \textit{editing} can enhance the image generation experience by supporting interactive refinement~\cite{ruiz2022dreambooth,imagic,unitune,hertz2022prompt}.
We focus on 
\task, where a user provides an image, a masked area, and a text prompt and the model fills the masked area, consistent with both the prompt and the image context (Fig.~\ref{fig:concept}).
This complements mask-free editing \cite{hertz2022prompt,unitune,imagic} with the precision of localized edits~\cite{nichol-glide,diffedit}.



\begin{figure}[t]
    \centering
    \includegraphics[width=\linewidth]{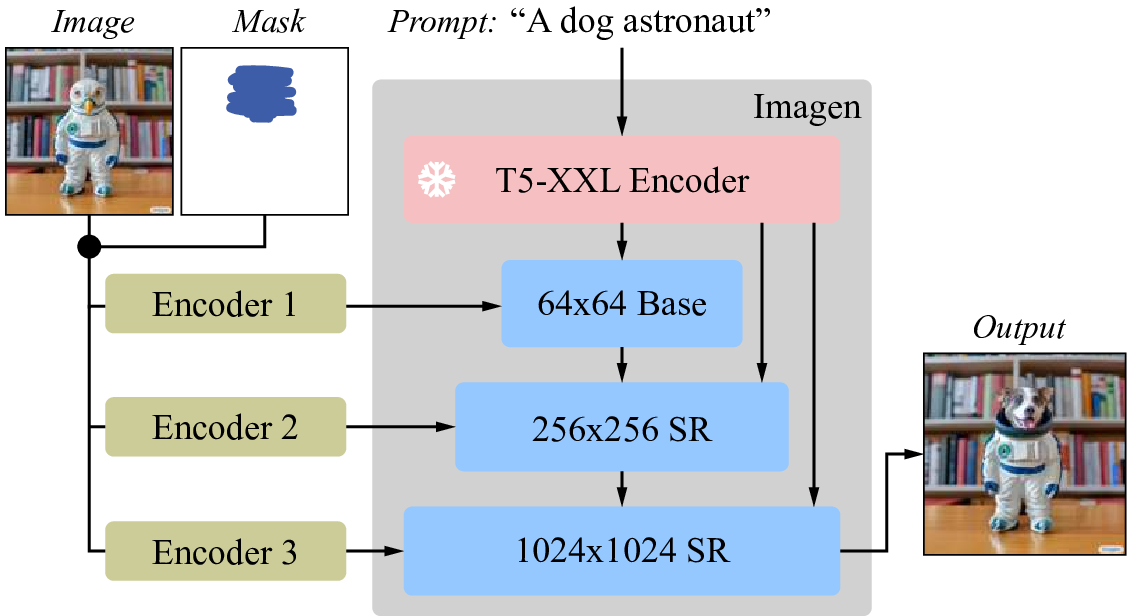}
    \caption{\textbf{Imagenator} is an image editing model built by finetuning Imagen. All of the diffusion models, i.e., the base model and super-resolution (SR) models, condition on high-resolution 1024$\times$1024 image and mask inputs. To this
    end, new convolutional image encoders are introduced.}
    \vspace{-2mm}
    \label{fig:imagenatorunet}
\end{figure}

This paper contributes to the modeling and evaluation of text-guided image inpainting.
Our modeling contribution is \bfit{\imagenator{}},\footnote{\url{https://imagen.research.google/editor/}}
a text-guided image editor that combines large scale language representations with fine-grained control to produce high fidelity outputs.
\imagenator{} is a cascaded diffusion model that extends Imagen~\cite{sahariac-imagen} through finetuning for \task. \imagenator{} adds image and mask context to each diffusion stage via three convolutional downsampling image encoders, shown in Fig.~\ref{fig:imagenatorunet}.

\begin{figure}[t]
    \centering
    \includegraphics[width=\linewidth]{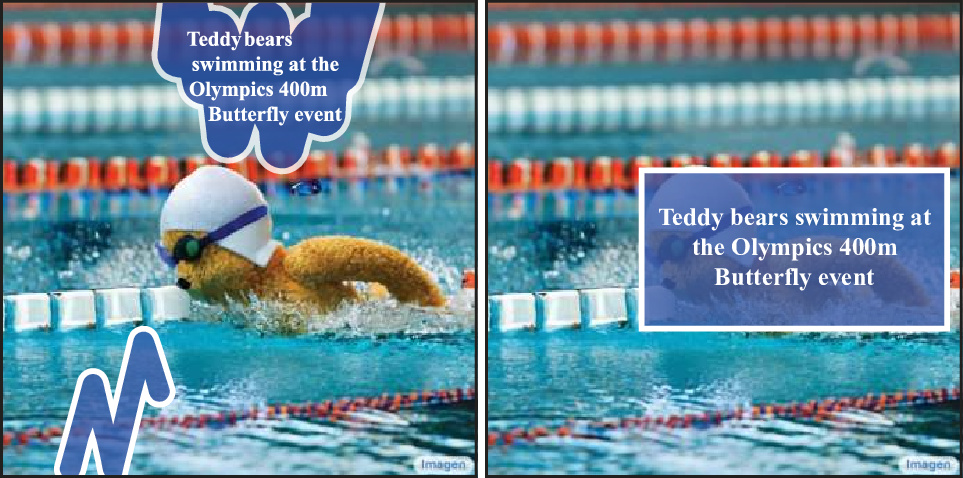}
    \caption{Random masks (left) frequently capture background or intersect object boundaries, defining regions that can be plausibly inpainted just from image context alone. \textit{Object masks} (right) are harder to inpaint from image context alone, encouraging models to rely more on text inputs during training. (Note: This example image was generated by Imagen and is not in the training data.)}
    \label{fig:masking_strategies}
\end{figure}

A key challenge in \task is ensuring that generated outputs are faithful to the text prompts. 
The standard training procedure uses randomly masked regions of input images~\cite{sahariac-palette,nichol-glide}. We hypothesize that this leads to weak image-text alignment since randomly chosen regions can often be plausibly inpainted using only the image context, without much attention to the prompt. 
We instead propose a novel \textit{object masking} technique that encourages the model to rely more on the text prompt during training (Fig.~\ref{fig:masking_strategies}). This helps make \imagenator{} more controllable and substantially improves text-image alignment.

Observing that there are no carefully-designed standard datasets for evaluating \task, we propose
\bfit{\eb{}}, a 
curated evaluation dataset that captures a wide variety of language, types of images, and levels of difficulty. Each \eb{} example consists of (i) a masked input image, (ii) an input text prompt, and (iii) a high-quality output image that can be used as reference for automatic metrics. To provide insight into the relative strengths and weaknesses of different models, edit prompts are categorized along three axes: attributes (material, color, shape, size, count), objects (common, rare, text rendering), and scenes (indoor, outdoor, realistic, paintings). 

Finally, we perform extensive human evaluations on \eb{},
probing \imagenator{} alongside \stablediffusion{} (\sd{})~\cite{rombach-cvpr-2022}, and \dalle{} (\dl{})~\cite{ramesh-dalle2}. 
Human annotators are asked to judge a) \emph{text-image alignment} -- how well the prompt is realized (both overall and assessing the presence of each object/attribute individually) and b) \emph{image quality} -- visual quality regardless of the text prompt.
In terms of text-image alignment, \imagenator{} trained with object-masking is preferred in 68\% of comparisons with its counterpart configuration trained with random masking (a commonly adopted method \cite{lama-suvorov2021resolution,sahariac-palette,nichol-glide}). Improvements are across-the-board in all object and attribute categories. \imagenator{} is also preferred by human annotators relative to \sd{} and \dl{} (78\% and 77\% respectively). As a cohort, models are better at object-rendering than text-rendering, and handle material/color/size attributes better than count/shape attributes. Comparing automatic evaluation metrics with human judgments, we conclude that while human evaluation remains indispensable, CLIPScore~\cite{hessel2021clipscore} is the most useful metric for hyperparameter tuning and model selection. 

In summary, our main contributions are: (i) \imagenator{}, a new state-of-the-art diffusion model for high fidelity text-guided image editing 
(Sec.~\ref{sec:imagenator}); (ii) \eb{}, a manually curated evaluation benchmark for \task{} that assesses fine-grained details such as object-attribute combinations (Sec.~\ref{sec:eb}); and (iii) a comprehensive human evaluation on \eb{}, highlighting the relative strengths and weaknesses of current models, and the usefulness of various automated evaluation metrics for text-guided image editing (Sec.~\ref{sec:evaluation}).

\section{Related Work}

\noindent
\textbf{Text-Guided Image Editing.} 
There has been much recent work on \task \cite{paintbyword,kim-emnlp-2016,blendeddiffusion,diffedit,nichol-glide,ramesh-dalle2,ding2022cogview2,rombach-cvpr-2022}. Paint By Word~\cite{paintbyword} optimizes for a balance between a) the consistency between the input and edited images, and b) the consistency between the text guide and the edited image. The technique has been used effectively more recently in DiffusionCLIP~\cite{kim2021diffusionclip}. Blended Diffusion~\cite{blendeddiffusion} runs CLIP-guided diffusion on the foreground (masked region) and the background (the context) in parallel and separately, and then blends the result by element-wise aggregation. CogView2~\cite{ding2022cogview2} proposes an auto-regressive text-guided infilling technique powered by cross-modal language modeling. DiffEdit~\cite{diffedit} presents a ``masked mask-free'' formulation where masking segmentation and masked diffusion are run in parallel to apply masked inpainting. Most relevant to our work are Stable Diffusion~\cite{rombach-cvpr-2022} and GLIDE/DALL-E 2~\cite{nichol-glide,ramesh-dalle2}, which are also diffusion models. Key differences in our work are the use of an object detector for masking plus architectural changes to enable high resolution editing. 



There has also been much work in \bfit{mask-free} text-guided image editing \cite{text2live,hertz2022prompt,imagic,unitune}. Text2Live~\cite{text2live} operates on an isolated edit-layer with semantic localization, which allows for good context preservation yet does not lend itself well to extensive modifications. Prompt-to-Prompt~\cite{hertz2022prompt} presents powerful manipulation techniques on the cross-attention in the text-conditioning module. Imagic~\cite{imagic} optimizes a special embedding to capture the semantics of the input image, and produces textually faithful edits by interpolating the optimized embedding with the embedding of the target text. 
\noindent\textbf{Evaluation of Text-Guided Image Editing.}
Text-guided image inpainting has primarily been evaluated with respect to three aspects. (1) Image quality \cite{murray-12,nichol-glide,blendeddiffusion,manitrans} assesses the standalone quality of an image, usually independent of a (single) ground truth reference. (2) Reconstruction fidelity \cite{manigan,kim2021diffusionclip,gao-ying-22}, on the other hand, calculates a similarity between the evaluated image and \emph{a} ground truth. (3) Text-image alignment \cite{park2021benchmark,de-net,zhou-yufan-22} measures similarity between visual outputs and textual inputs. (1) and (3) are most relevant to our work because \task promotes diverse coverage (contingent on semantic coherence) rather than faithfulness to one particular reference.

\bfit{Automatic evaluation.} The standard automatic metric for image quality is \emph{Frech\'et Inception Distance (FID)}, which assesses the quality of images in the latent space of a generative model with respect to the distribution of a set of real images. For text-image alignment, metrics based on text-image encoders (notably CLIP \cite{clip-paper}) have been popular, e.g. CLIPScore~\cite{hessel2021clipscore} - distance between text and image encodings; CLIP-R-Precision~\cite{park2021benchmark} - the retrieval rank of the edited/synthesized image for the ground truth text among distractors. In this work, we further explore the connection between automatic and human evaluation, gauging the extent to which the automatic metrics agree with human assessments of model performance (for which there is currently no substitute when judging model outputs).


\bfit{Human evaluation.} The most typical formulation is asking two questions about side-by-side outputs from competing models -- \emph{which has the better image quality?}, and \emph{which aligns with this \$\{\texttt{text}\} better?} (paraphrased in varied ways). \eb{} extends this paradigm by constructing a benchmark along diverse feature axes (attribute/object/scene), focusing the evaluation on the masked area rather than the full image (which delineates the evaluation of image \emph{editing} from \emph{generation}), and asking annotators to assess the presence of each object and attribute mentioned in the prompt \textit{individually}. This distinguishes our work from the previous efforts in \task,
which typically evaluate in a less systematic manner or merely share cherry-picked examples~\cite{nichol-glide,zhou-yufan-22,kim2021diffusionclip,blendeddiffusion}.

\section{Imagen Editor}
\label{sec:imagenator}

\noindent
\imagenator{} is a \tasklong model targeting 
improved representation and reflection of linguistic inputs,
fine-grained control and high fidelity outputs. \imagenator{} takes three inputs from the user, 1) the image to be edited, 2) a binary mask to specify the edit region, and 3) a text prompt -- and all three inputs are used to guide the output samples. \imagenator{} is a diffusion-based model \cite{de2021diffusion} fine-tuned from Imagen \cite{sahariac-imagen} for editing.  See Figure \ref{fig:imagenatorunet} for an illustration. 


\noindent
\textbf{Object Detector Masking Policy.} 
A natural question to ask is: what kind of masks do we use to train models for \tasklong{}? The masked regions should be well aligned to the edit text prompt. Ideally, we would have a large expert dataset of aligned mask-prompt edits to train on; however, such a dataset does not exist and curating a large one would be difficult. 
One natural, simple policy to use is a random mask distribution, for example random box and/or random stroke masks; this has been successfully applied to prior inpainting models \cite{yu2018generative, yu2019free, sahariac-palette}. However, when random masks are used during training, they may cover a region irrelevant to the text prompt (Fig~\ref{fig:masking_strategies} left). 
Training on such examples can encourage the model to ignore the text prompt.
We find this issue to be especially prevalent when masked regions are small or only partially cover an object, which was similarly observed for CogView2~\cite{ding2022cogview2}.

Unlike simple text-\textbf{un}conditional inpainting, we need generated regions (from the mask) to not only be realistic, but also to relate coherently to the input text prompt. We propose a simple, effective solution to this problem. We hypothesize that masking out identified \textit{objects} entirely will induce a greater overlap with the text prompt (Fig~\ref{fig:masking_strategies}), and consequently encourage the model to pay more attention to the text prompt when inpainting. We use an off-the-shelf object detector to detect and localize objects, and use these bounding object boxes to generate masks to be used during training.  The model we use is the lightweight SSD Mobilenet v2~\cite{ssd-mobilenet-v2}  \footnote{\scriptsize{\url{https://tfhub.dev/tensorflow/ssd_mobilenet_v2/2}}} which can be easily run on-the-fly and thus offers the same flexibility as random masking policies. Our experiments show that this simple modification to the masking policy works surprisingly well, and it alleviates most of the issues faced by models trained with a random masking policy.
See the Appendix
for implementation details.

\noindent
\textbf{High-Resolution Editing.} 
In \imagenator{}, we modify Imagen to condition on both the image and the mask by concatenating them with the diffusion latents along the channel dimension, similar to SR3~\cite{saharia2021image}, Palette~\cite{sahariac-palette} and GLIDE~\cite{nichol-glide}. The conditioning image and the corresponding mask input to \imagenator{} are always at 1024$\times$1024 resolution. The base diffusion 64$\times$64 model and the 64$\times$64$\rightarrow$256$\times$256 super-resolution model operate at a smaller resolution, and thus require some form of downsampling to match the diffusion latent resolution (e.g., 64$\times$64 or 256$\times$256). One method is to use a parameter-free downsampling operation (e.g., bicubic); we instead apply a parameterized downsampling convolution (e.g., convolution with a stride). In initial experiments we found this parameterized downsampling operation to be critical for high fidelity. Simple bicubic downsampling resulted in significant artifacts along the mask boundaries in the final output image, and switching to a parameterized downsampling convolution resulted in much higher fidelity. 
We also initialize the corresponding new input channel weights to zero (like \cite{nichol-glide}); this means that at initialization the model is identical to Imagen, as it ignores the conditioning image and mask.


\noindent
\textbf{Classifier-Free Guidance.}
Classifier-Free Guidance (CFG) \cite{ho2021classifierfree} is a technique to bias samples to a particular conditioning (e.g., text prompt), at the cost of mode coverage. CFG has been found to be highly effective in boosting text-image alignment as well as image fidelity in text$\rightarrow$image models \cite{nichol-glide,sahariac-imagen,gafni2022make,yu-parti}. We found CFG continues to be critical for ensuring strong alignment between the generated image and the input text prompt for \tasklong.
We follow \cite{ho2022imagen} and use high guidance weights with guidance oscillation. In the base model, where ensuring strong alignment with text is most critical, we use a guidance weight schedule which oscillates between 1 and 30. We observe that high guidance weights combined with oscillating guidance \cite{ho2022imagen} result in the best trade-off between sample fidelity and text-image alignment.
\section{\eb{}}
\label{sec:eb}

\begin{figure}
    \centering
    \includegraphics[width=0.83\linewidth]{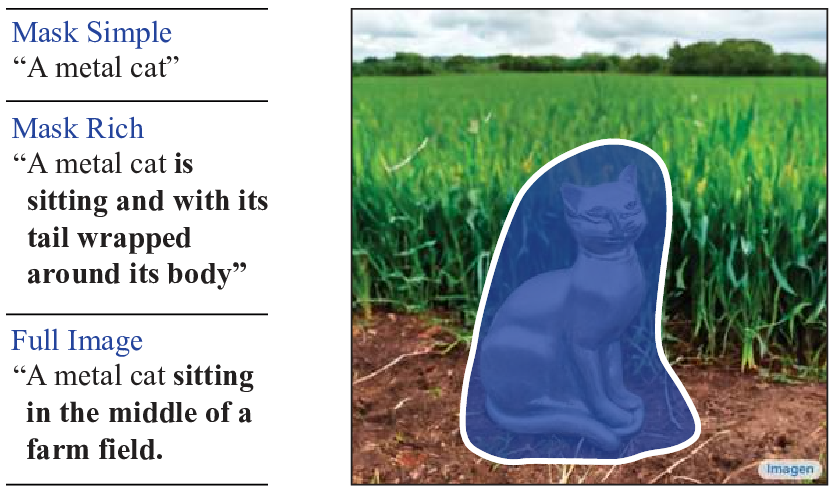}
    \caption{\textbf{\eb{} example}. The full image is used as a reference for successful inpainting. The mask covers the target object with a free-form, non-hinting shape. The three descriptions types are: single-attribute description of the masked object (\bfit{\simple}), multi-attribute description of the masked object (\bfit{\rich}), or whole image (\bfit{\full}). \rich especially probes models' ability to handle complex attribute binding and inclusion \cite{binding-theory}.
    }
    \label{fig:eb_item_example}
\end{figure}

\noindent
\textbf{Overview.} \eb{} is a new benchmark for \task{} based on 240 images. Each image is paired with a mask that specifies the image region to be modified via inpainting. For each image-mask pair, we provide three different text prompts, representing different approaches to specifying the edit (see Fig. \ref{fig:eb_item_example}). Similar to the DrawBench~\cite{sahariac-imagen} and PartiPrompts~\cite{yu-parti} benchmarks for text-to-image generation, \eb{} is hand-curated to capture a wide variety of categories and aspects of difficulty.

\noindent
\textbf{Image Collection.}
\eb{} includes both natural images drawn from existing computer vision datasets (Visual Genome~\cite{krishnavisualgenome} and Open Images~\cite{OpenImages}), and synthetic images generated by text-to-image models (Imagen~\cite{sahariac-imagen} and Parti~\cite{yu-parti}) at 50:50 ratio.
To construct \eb{}, we first generate a wide variety of initial prompts to guide the image collection process. Initial
prompts are generated by enumerating \emph{attribute-object-scene} combinations from these categories:
\begin{compactitem}
    \item \emph{Attributes}: \emph{\{material, color, shape, size, count\}};
    \item \emph{Objects}: \emph{\{common, rare, text-rendering\}};
    \item \emph{Scenes}: \emph{\{indoor, outdoor, realistic, painting\}}.
\end{compactitem}
The choice of object, attribute and scene categories was inspired by studying image editing requests on Reddit.\footnote{\url{https://www.reddit.com/r/PhotoshopRequest/}}
Natural images are selected by manually searching for images matching object, attribute, scene combinations, e.g., for \aos{material}{common}{outdoor}, images of an outdoor patio made of wood could be selected, instantiating  \aos{wooden}{patio}{outdoor}. Synthetic images are created by sampling an object and attribute within each category (e.g., instantiating \aos{material}{common}{outdoor} as \aos{metal}{cat}{outdoor}), writing a matching prompt (e.g., \emph{a metal cat standing in the middle of a farm field.}), sampling batches of images from text-to-image models as candidates, and then manually identifying the generated image that best matches the prompt. As shown in Figure \ref{fig:eb_item_example}, synthetic images can capture object-attribute-scene combinations that are unlikely to occur naturally, and editing these images is an important use case as part of the workflow of image creation combining descriptions and gesture.

\begin{figure}
    \centering
    \includegraphics[width=\linewidth]{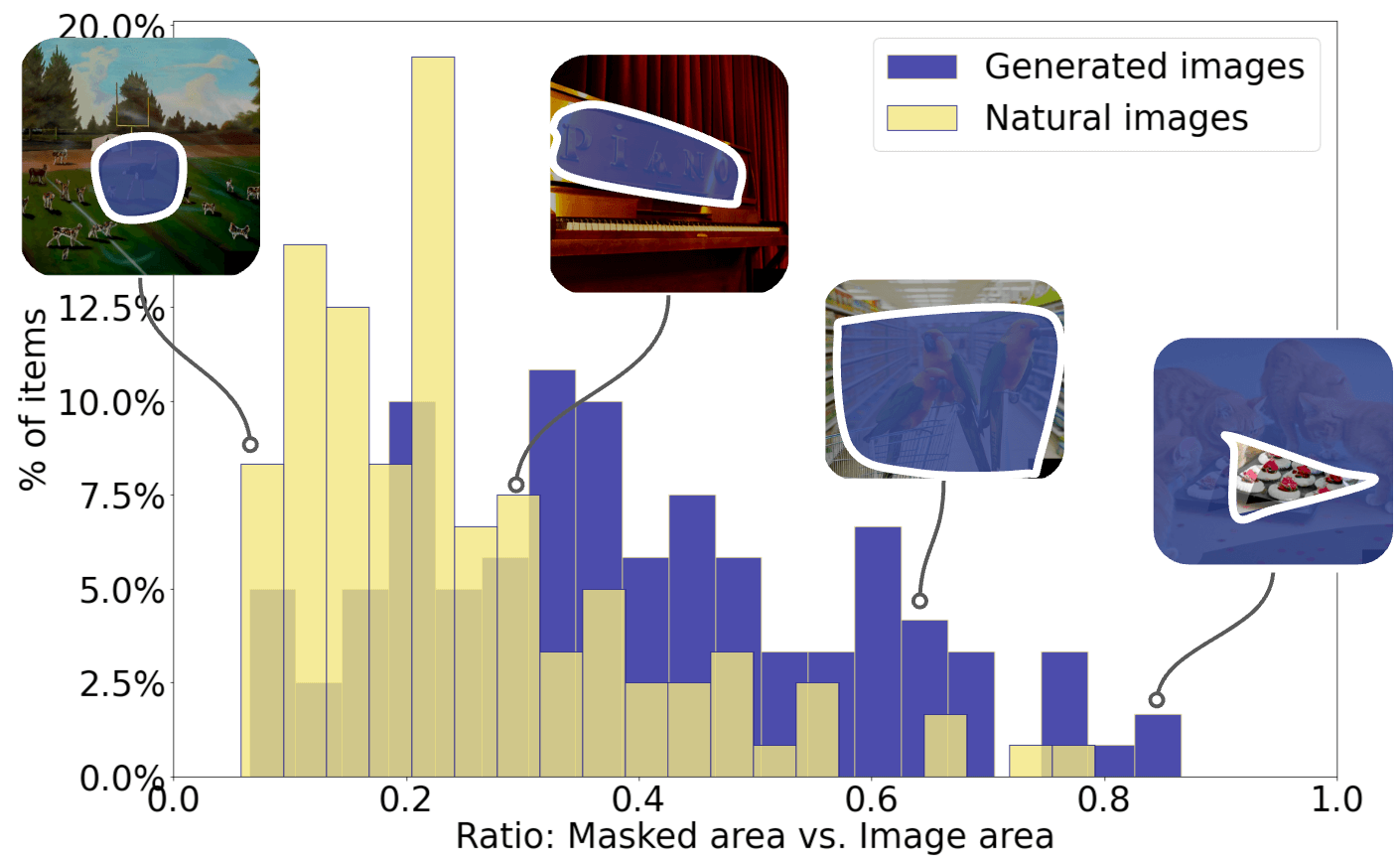}
    \caption{EditBench encompasses a wide variety of mask sizes, including large masks that contact the edges of the images (which can amount to an \textit{uncropping} task in some cases).}
    \label{fig:eb_mask_ratio_dist}
\end{figure}

\noindent
\textbf{Image Masks.} For each image, we manually-annotate a free-form mask that completely covers the target object. We are careful not to too-closely segment the target object, which could leak information about the object underneath via its shape. We also include masks with a range of sizes (Fig. \ref{fig:eb_mask_ratio_dist}) to check the undesirable sensitivity to mask sizes \cite{rodrigues-21}. We check models' robustness against the tendency of painting-over small masks due to overwhelming influence from the context; we also evaluate large-area inpainting and uncropping where the challenge is to not completely disregard the relatively small context.

\noindent
\textbf{Creating Text Prompts.}
Prior work (e.g., GLIDE~\cite{nichol-glide}) often demonstrate text-guided image inpainting using prompts that describe the full image. However, writing full-image descriptions is unnatural for the use case of inpainting particular components of an image that depicts a complex scene with multiple objects and characters. Consequently, for each image-mask pair, we create three text prompts to probe model behavior from different angles. One type of prompt gives only a basic description (\bfit{\simple}) for the mask, another gives much more details (\bfit{\rich}), and the last describes the full image (\bfit{\full}, disregarding the mask). The unmasked input image itself serves as a reference image for inpainting in accordance with the prompts. See Fig. \ref{fig:eb_item_example} for a summary.

\section{Evaluation}
\label{sec:evaluation}

\noindent
We conduct comprehensive human evaluations of both text-image alignment and image quality on \eb{}.
We also analyze human preferences relative to automatic metrics.
We evaluate four models:
\begin{compactitem}
    \item \textbf{\imagenator{} (\im{})}: Our full model described in Sec. \ref{sec:imagenator};
    \item \textbf{\imagenatorrm{} (\imrm{})}: \imagenator{} finetuned with \textbf{R}andom \textbf{M}asking instead of object masking;
    \item \textbf{\stablediffusion{} (\sd{})}: Version 1.5 of the model based on Rombach et al.~\cite{rombach-cvpr-2022};\footnote{\scriptsize{\url{http://huggingface.co/runwayml/stable-diffusion-v1-5}}}
    \item \textbf{\dalle{} (\dl{})}: A commercial web UI based on Ramesh et al.~\cite{ramesh-dalle2}, accessed in October 2022.\footnote{\scriptsize{\url{https://openai.com/dall-e-2/}}}
\end{compactitem}
\noindent
We compare \imagenator{} with \imagenatorrm{} to quantify the benefits of object masking during training. We include evaluations of \stablediffusion{} and \dalle{} to place our work in context with prior work and to more broadly analyze the limitations of the current state of the art.

\subsection{Human Evaluation Protocol}
\label{subsec:human-eval}

\begin{figure}[t]
    \centering
    \includegraphics[width=\linewidth]{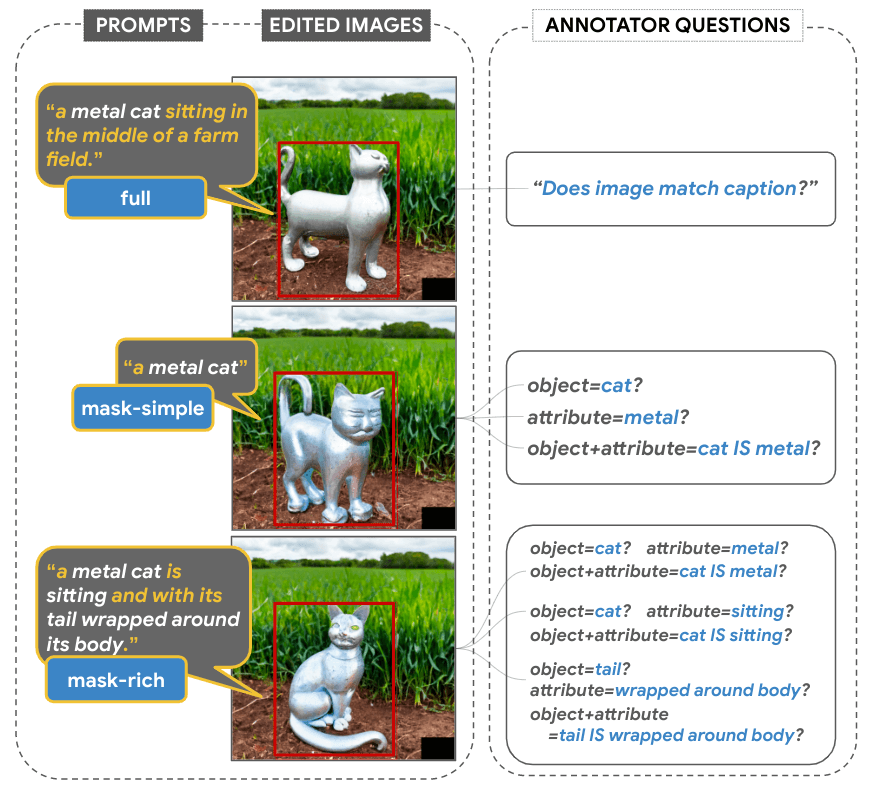}
    \caption{Human evaluation for single model text-image alignment. \full elicits annotators' overall impression of text-image alignment; \simple and \rich check for the correct inclusion of particular attributes and objects, and attribute binding.}
    \label{fig:human_eval_ui_text_alignment}
\end{figure}

\noindent
We perform two types of human evaluations: \textit{single image} evaluations and forced choice \textit{side-by-side image} evaluations. The former allow us to ask fine-grained questions to establish if each individual object and attribute in the prompt has been correctly rendered. Side-by-side evaluations focus on comparisons between \imagenator{} and the other models, capturing \textit{relative} model performance. We evaluate text-image alignment in both settings, and overall image quality only in side-by-side evaluations.
In all evaluations we use a red box to highlight the image region edited by the model and ask the annotator to pay special attention to it (Fig. \ref{fig:human_eval_ui_text_alignment}). Each model is evaluated based on four sampled image edits for each prompt.

\noindent
\textbf{Single Image Evaluations.}
Our single image evaluations are adapted to the given detail level for each type of prompt. For the \bfit{Full} prompts (describing the full image), annotators assess general text-image alignment by giving a binary answer to the question \emph{Does the image match the caption?}. For \bfit{Mask-Simple} prompts describing the masked region with one object and one attribute (e.g. \textit{a metal cat}), evaluations are more fine-grained (Fig. \ref{fig:human_eval_ui_text_alignment}). Annotators answer three binary questions, evaluating: (1) whether the object (\textit{cat}) is rendered, (2) whether the given attribute (\textit{metal}) is present in the image, and (3) whether the attribute (\textit{metal}) is depicted applied to the correct object (\textit{cat})~\cite{yu-parti,binding-theory}. Finally, for \bfit{Mask-Rich} prompts, we extend the previous evaluation to multiple attribute-object pairs. The annotator answers three sets of three binary questions -- about the attribute, the object, and the attribute binding -- making 9 binary judgments in total. Compared to previous evaluations in image generation that only assess general text-image alignment~\cite{nichol-glide,zhou-yufan-22,blendeddiffusion}, our fine-grained evaluations provide greater insight into language fidelity and also which categories of objects and attributes present the most difficulties. Annotators in total perform 11.5K single model evaluation tasks (240 images $\times$ 3 prompts $\times$ 4 models $\times$ 4 samples).

\noindent
\textbf{Side-by-Side Evaluations.}
Finally, for \bfit{Mask-Rich} prompts, we extend the previous evaluation to multiple attribute-object pairs. The annotator answers three sets of three binary questions -- about the attribute, the object, and the attribute binding -- making 9 binary judgments in total. Compared to previous evaluations in image generation that only assess general text-image alignment~\cite{nichol-glide,zhou-yufan-22,blendeddiffusion}, our fine-grained evaluations provide greater insight into language fidelity and also which categories of objects and attributes present the most difficulties. 18 (US-based) annotators in total perform 11.5K single model evaluation tasks (240 images $\times$ 3 prompts $\times$ 4 models $\times$ 4 samples).
We describe the human evaluation process in more detail in the Appendix.

\subsection{Human Evaluation Results}
\label{subsec:human-eval-results}

\begin{figure}
    \centering
    \includegraphics[width=\linewidth]{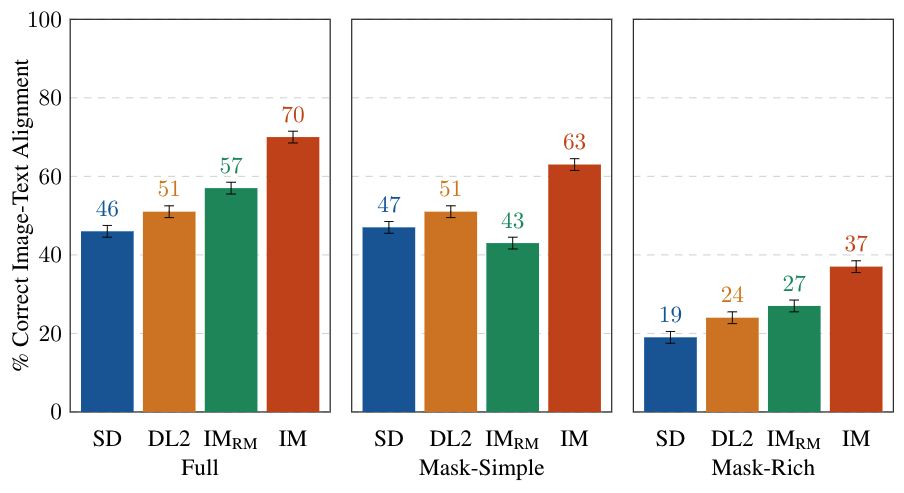}
    \vspace{-6mm}
    \caption{Single-image human evaluations of \tasklong on \eb{} by \textit{prompt type}. In this figure, for Mask-Simple and Mask-Rich prompts, text-image alignment is only counted as correct if the edited image correctly includes \textit{every} attribute and object specified in the prompt, including the correct attribute binding (setting a very high bar for correctness).  Note that due to different evaluation designs, Full vs Mask-only prompts results are less directly comparable.
    }
    \label{fig:heval-overall-1}
\end{figure}


\begin{figure}
    \centering
    \includegraphics[width=1\linewidth]{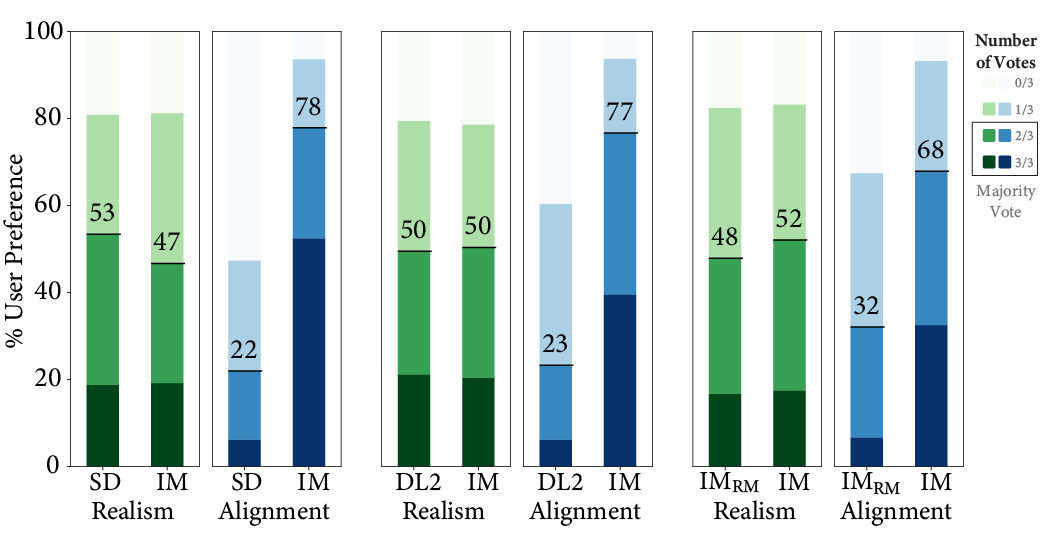}
    \caption{Side-by-side human evaluation of image realism \& text-image alignment on \eb{} Mask-Rich prompts. For text-image alignment, \imagenator{} is preferred in all comparisons.}
    \label{fig:heval-overall-2}
\end{figure}

\begin{figure}
    \centering
    \includegraphics[width=\linewidth]{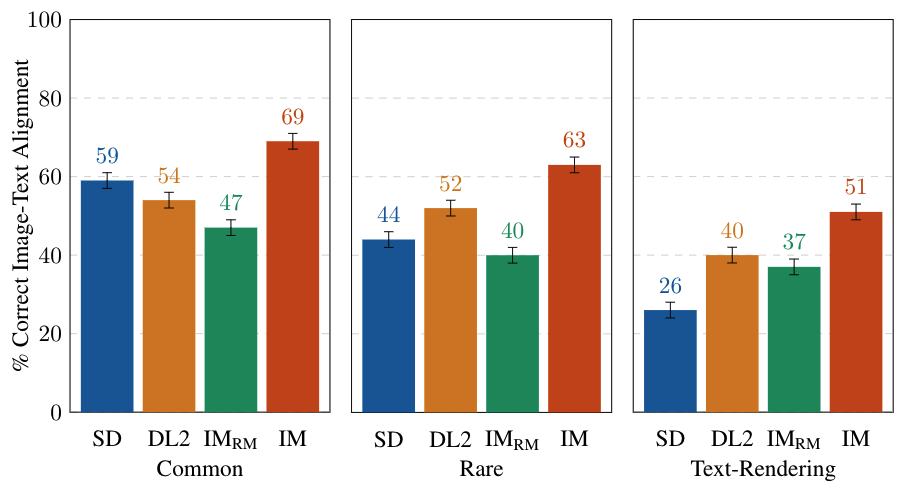}
    \vspace{-6mm}
    \caption{Single-image human evaluations on \eb{} Mask-Simple by \emph{object type}. As a cohort, models are better at object-rendering than text-rendering.}
    \label{fig:heval-breakdown-1}
\end{figure}

\begin{figure}
    \centering
    \includegraphics[width=\linewidth]{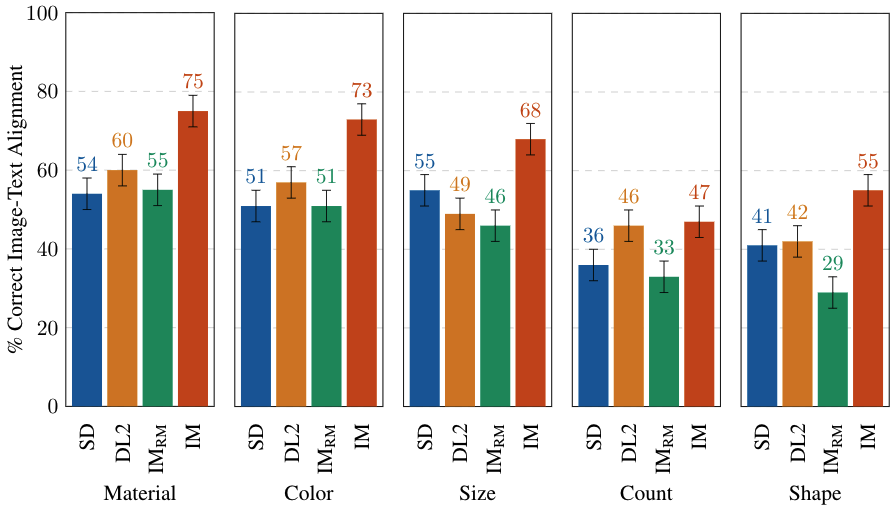}
    \vspace{-6mm}
    \caption{Single-image human evaluations on \eb{} Mask-Simple by \emph{attribute type}. Object masking improves adherence to prompt attributes across-the-board (\im{} vs. \imrm{}).}
    \label{fig:heval-breakdown-2}
\end{figure}


\noindent
\textbf{Overall.} Fig. \ref{fig:heval-overall-1} presents the aggregated human ratings, sliced by prompt types. \% Correct Image-Text Alignment is the proportion of positive judgments a model receives. Each question is binary -- in the case of \full, responses reflect the overall impression, whereas for \simple and \rich a positive response indicates the edited image attributes are correctly bound to the correct objects. Across-the-board, \imagenator{} receives the highest ratings (10-13\% higher than the 2nd highest). For the rest, the performance order is \imrm $>$ \dl $>$ \sd (with 3-6\% difference) except for with \simple, where \imrm falls 4-8\% behind. As relatively more semantic content is involved in \full and \rich, we conjecture \imrm and \im are benefited by the higher performing T5 XXL text encoder (see \cite{hertz2022prompt}, D1).

An interesting observation is annotators rate models higher with \full than \simple prompts, even though the former involves more semantic content. There are two likely reasons for this: a) models are trained by conditioning on \full prompts rather than mask-only \cite{imagen-paper,rombach-cvpr-2022,ramesh-dalle2}; b) since we do not change the context (unmasked), \full gains an advantage of having correct associations for this portion because the unmasked (and correct) pixels remain unchanged.

\begin{figure}
    \centering
    \includegraphics[width=\linewidth]{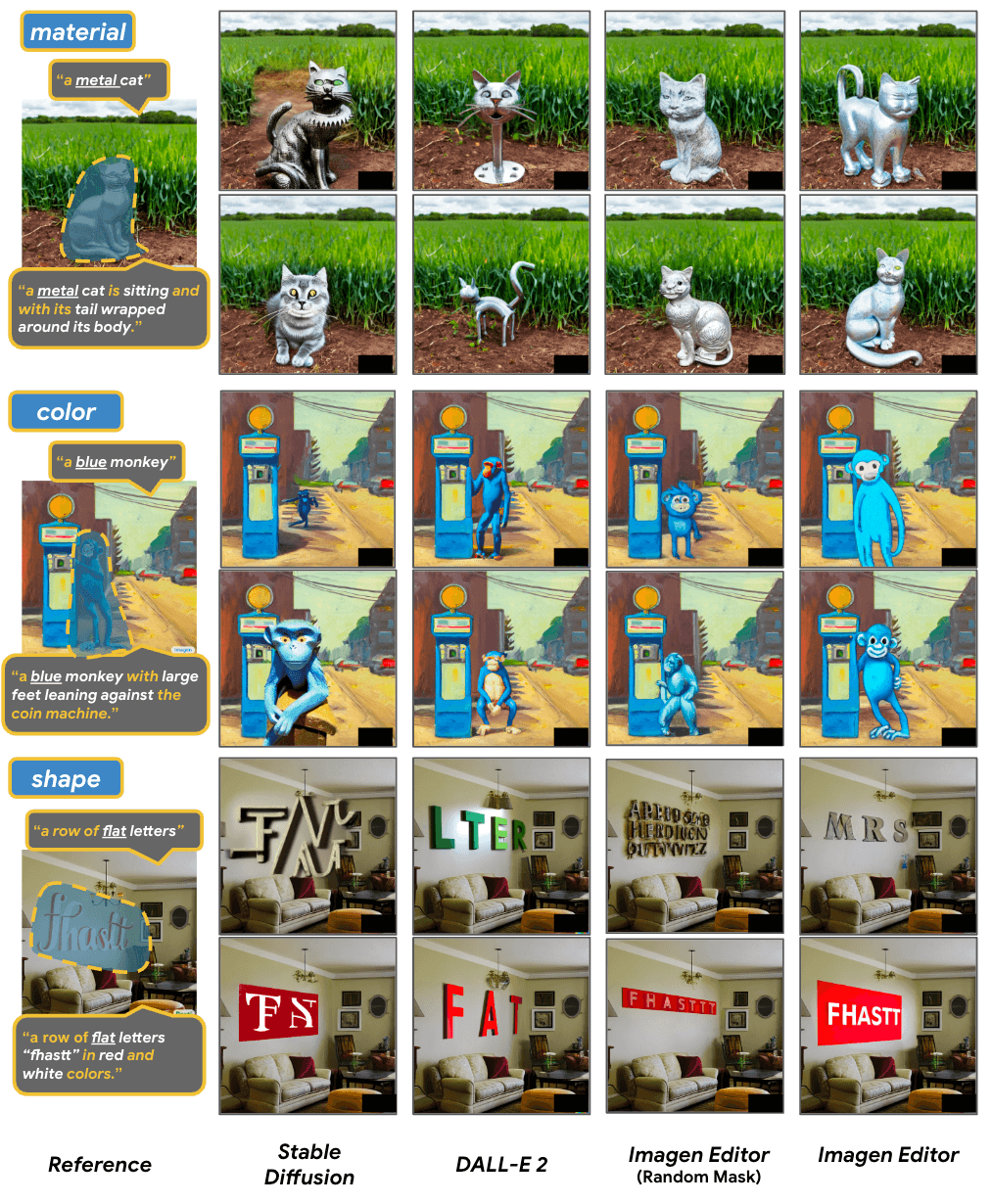}
    \caption{Example model outputs for \simple vs. \rich prompts. Object masking improves \imagenator{}'s fine-grained adherence to the prompt compared to the same model trained with random masking. See Appendix for more examples.}
    \label{fig:simple-vs-rich-splash}
\end{figure}

Finally, note that with \rich, while \im retains 10+\% lead over the rest 
(see Fig. \ref{fig:heval-overall-1} and examples in Fig. \ref{fig:simple-vs-rich-splash}),
the overall performance drops substantially -- leaving considerable room for future improvement. 

\noindent
\textbf{Side-by-Side.} In Fig. \ref{fig:heval-overall-2}, compared with other models 1v1, \im leads in text alignment with a substantial margin, being preferred by annotators 78\%, 77\%, and 68\% of the time compared to \sd, \dl, and \imrm respectively.  These gains were realized while achieving similar levels of performance (0-6\% delta) in image quality.

\noindent
\textbf{Breakdown by \emph{objects}.} In Fig. \ref{fig:heval-breakdown-1}, \im leads in all object types: 10\%, 11\%, and 11\% higher than the 2nd highest in \emph{common}, \emph{rare}, and \emph{text-rendering}. For the rest, the notable observation is \sd's performance plummets in text-rendering (59\% \& 44\% for \emph{common} and \emph{rare}, and only 26\% for \emph{text-rendering}).


\noindent
\textbf{Breakdown by \emph{attributes}.} In Fig. \ref{fig:heval-breakdown-2}, \im is rated much higher (13-16\%) than the 2nd highest, except for in \emph{count}, where \dl is merely 1\% behind. \im also improves the least over \imrm in this attribute type (14\%, with the 2nd lowest 22\%), making \emph{count} a particularly interesting category for future study. In general models get rated lower in \emph{count} and \emph{shape}, and the two happen to be the more abstract. The result is intuitive yet runs counter to findings in the recognition of abstract attributes \cite{nan-zhixiong-19}.
The relatively similar performance in \emph{size} vs. \emph{material}/\emph{count} is slightly unexpected -- it is a relational category where the understanding requires appropriate contextualization, while for the latter the object itself is the only source of information wrt. other objects or the general surrounding. One possibility is the simplicity of our design: we include straightforward comparison of different sizes but do not build adversarial cases where seemingly small/large objects are in actuality the opposite.

\subsection{Automatic Evaluation Metrics}
\label{subsec:auto-eval}

Although human evaluations are widely adopted as the gold standard for image realism and text-image alignment evaluations, automatic evaluation metrics are valuable for iterative hyperparameter tuning and model selection. We compare human judgments with automatic metrics to identify the best metrics for model development.


\noindent\textbf{Metrics.}
We investigate text-image alignment metrics based on \textbf{CLIPScore}~\cite{hessel2021clipscore} and \textbf{CLIP-R-Prec(ision)}~\cite{park2021benchmark}. CLIPScore calculates text-to-image (T2I) or image-to-image (I2I) similarity in the latent space of the contrastively-trained CLIP model~\cite{radford2021learning}.
CLIP-R-Prec is a ranking based approach (typically formulated as text R-Precision \cite{park2021benchmark}) that measures how well the generated image retrieves the text prompt with CLIP from among a set of text distractors. 
As text distractors we use all the other prompts in \eb{} with the same prompt type.

\noindent\textbf{Comparison to human judgments.}
To evaluate agreement between automatic metrics and human scores, a common practice is to report correlation coefficients~\cite{specia-etal-2021-findings}. However, metrics such as Spearman's $\rho$ consider the ranking induced by each score over \textit{all pairs} of observations, which includes ranking images with \textit{different prompts}. Rather than comparing images with different prompts, which is a difficult judgment even for people, we focus on two questions: (1) For a given prompt, can automatic metrics pick the image preferred by people? and (2) Can automatic metrics identify the model with the highest human evaluations?

\begin{table}[t]
\setlength{\tabcolsep}{3.0pt}
\begin{center}
\small
\begin{tabularx}{\linewidth}{Xlccccc}
\textbf{Prompt} & \textbf{Image} & \textbf{T2I} & \textbf{I2I} & \textbf{T2I+I2I} & \textbf{R-Prec} & \textbf{Rand} \\
\toprule
Full            & Full            & \textbf{70.1}      & 58.6               & 66.8                   & 53.3                 & 50.0            \\
Full            & Crop        & 68.1               & 55.8               & 62.4                   & 57.7                & 50.0            \\
\midrule
Mask-Simple     & Full            & 73.8               & 53.1               & 63.2                   & 72.0                 & 50.0            \\
Mask-Simple     & Crop        & \textbf{76.0}      & 55.3               & 66.4                   & 71.0                 & 50.0            \\
\midrule
Mask-Rich       & Full            & 66.7               & 55.2               & 63.4                   & 62.3                 & 50.0            \\
Mask-Rich       & Crop        & \textbf{68.4}      & 56.4               & 64.1                   & 63.3                 & 50.0            \\
\bottomrule          
\end{tabularx}
\caption{Percentage agreement between CLIPScore metrics and human judgments when picking the \textit{best image out of two model-generated images} for the same text prompt. Text-to-image (T2I) CLIPScore similarity outperforms CLIP-R-precision (R-Prec) and image-to-image (I2I) similarity using a reference image. }
\label{tab:image-level-results}
\end{center}
\end{table}

\begin{table}[t]
\setlength{\tabcolsep}{3.5pt}
\begin{center}
\small
\begin{tabularx}{\linewidth}{Xlccccc}
\textbf{Prompt} & \textbf{Image} & \textbf{T2I} & \textbf{I2I} & \textbf{T2I+I2I} & \textbf{R-Prec} & \textbf{Rand} \\
\toprule
Full            & Full            & \textbf{38.5}      & 30.8               & 35.7                   & 28.7                 & 25.0 \\
Full            & Crop        & 36.7               & 28.2               & 32.4                   & 27.9                 & 25.0            \\
\midrule
Mask-Simple     & Full            & 45.7               & 28.2               & 37.1                   & 40.2                 & 25.0            \\
Mask-Simple     & Crop       & \textbf{47.1}      & 29.4               & 38.5                   & 39.7                 & 25.0            \\
\midrule
Mask-Rich       & Full            & 45.8               & 31.4               & 40.5                   & 39.1                 & 25.0            \\
Mask-Rich       & Crop        & \textbf{48.1}      & 30.9               & 40.9                   & 39.3                 & 25.0           \\
\bottomrule          
\end{tabularx}
\caption{Agreement between CLIPScore metrics and human judgments when repeatedly picking the \textit{best model out of four hybrid models}. Text-to-image (T2I) similarity outperforms CLIP-R-precision (R-Prec) and reference image-to-image (I2I) similarity. }
\label{tab:model-level-results}
\end{center}
\vspace{-2mm}
\end{table}

In Tab. \ref{tab:image-level-results} we report the agreement between various metrics based on CLIPScore and human judgments when picking the \textit{best image} from two model-generated images with the same text prompt. We sample 10K image pairs and the best image in each pair is determined by human single-image evaluation scores (image pairs with the same human score are excluded). Metrics are calculated using both the full image (Full) and a cropped bounding box around the masked region (Crop). We find that CLIPScore based on text-to-image (T2I) similarity has the highest agreement with human judgments, identifying the best image in 68-76\% of pairs, depending on the prompt type. Unsurprisingly, CLIPScore has higher agreement with humans on simpler prompts (Mask-Simple) compared to more complex prompts (Mask-Rich).

In Tab. \ref{tab:model-level-results} we report agreement with human judgments when picking the \textit{best model} based on evaluations aggregated across \eb{}. Each evaluation is a choice between 4 hybrid models created by randomly selecting one sampled image from one of the available models for each prompt~\cite{graham-liu-2016-achieving,bert-score}. Scores for each hybrid model are created by averaging the scores for the corresponding images in the sampled data, and 100K evaluations are performed. Similarly to Tab. \ref{tab:image-level-results}, we find that CLIPScore (T2I) is most reliable metric (identifying the best hybrid model out of 4 in 39-48\% of instances). In both experiments CLIPScore works best when the image region matches the prompt, i.e., full image (Full) when the prompt describes the full image, and cropped bounding box around the masked region (Crop) when the text prompt describes only the masked region. In both Tab. \ref{tab:image-level-results} and Tab. \ref{tab:model-level-results} the 95\% confidence intervals calculated with bootstrap resampling are below 1\%.

\begin{table}[t]
\setlength{\tabcolsep}{6pt}
\begin{center}
\small
\begin{tabularx}{\linewidth}{Xcccc|c}
 & \textbf{\sd} & \textbf{\dl} & \textbf{\imrm} & \textbf{\im} & \textbf{Ref.} \\
 \toprule
 CLIPScore ($\uparrow$) & & & & \\
 \phantom{,,,,}T2I &    29.7 &   29.1 &   29.6 &        \textbf{31.5} &      31.0 \\ 
 \phantom{,,,,}I2I &    74.9 &   76.1 &   75.8 &        \textbf{76.6} &       -    \\
 \phantom{,,,,}T2I+I2I &    52.3 &    52.6 &    53.1 &        \textbf{53.6} &       -    \\
 CLIP-R-Prec ($\uparrow$) &  96.5 &  95.3 &  95.0 &        \textbf{98.6} &     \fbox{99.3} \\   
 NIMA ($\uparrow$)&   4.44 &   4.33 &   4.56 &        \textbf{4.63} &   \fbox{4.89} \\
 \bottomrule
\end{tabularx}
\caption{Aggregated automated metric scores. \textbf{boldface}: highest scoring model; \fbox{box}: reference images rated highest.}
\label{tab:t2i-overall}
\end{center}
\end{table}


\paragraph{Overall results.}
In Tab. \ref{tab:t2i-overall} we report automatic metrics aggregated over all prompts for each model, and for the \eb{} reference images. For CLIPScore metrics, the image representation (Full or Crop) is aligned with the prompt (Full or Mask). We also report \textbf{NIMA}~\cite{talebi-21} -- a model-based perceptual image quality metric. We find that the reference images receive the highest CLIP-R-Precision, and \imagenator{} is ranked highest among the 4 models on each metric.

\section{Societal Impact}
\label{sec:societal_impact}

\noindent
The image editing models presented are part of the growing family of generative models which unlock new capabilities in content creation, however, they also have the potential to create content that is harmful to individuals or to society.  In language modeling, it is now well recognized \cite{lambda,tan-neurips-2019} that text generation models are prone to recapitulating and amplifying social biases that may be present in their training sets.  The risk of amplifying social harm also pertains to text-to-image generation and \task; as discussed elsewhere, the data used to train these models is equally fraught \cite{sahariac-imagen, yu-parti, birhane2021laionaudit}.

A particular risk that is exposed by \task, but is not present in text-to-image models is that inpainting might enable the scaled and simple creation of convincing misinformation-- for example, editing an image of a political figure to include a controlled substance.  Two approaches to mitigating this risk that we have taken in our experimentation thus far are to (1) ensuring that distinctive watermarks are present on each generated image, and (2) refraining from photorealistic generation of human faces.
To extend the protections against misinformation, robust methods for proving image provenance such as steganographic watermarking are helpful \cite{leca-watermarking}. 
In addition, de-duplication of text-image training datasets can reduce the likelihood that a model reproduces a training set image \cite{lee-dedup}.\footnote{\url{https://openai.com/blog/dall-e-2-pre-training-mitigations/}}
That said, robust guardrails are needed to prevent recognizable likenesses of people from being generated and exposed to users.
\section{Conclusion}
We presented \imagenator{} and \eb{}, making significant advancements in \tasklong and the evaluation thereof. \imagenator{} is a \tasklong finetuned from Imagen. Key to \imagenator{} is adding new convolution layers to enable high-resolution editing, and the use of a object masking policy for training. \eb{} is a comprehensive systematic benchmark for \tasklong. \eb{} systematically evaluates \tasklong across multiple dimensions: attributes, objects, and scenes. We find \imagenator{} to outperform DALL-E 2 and Stable Diffusion on \eb{} in both human evaluation and automatic metrics.



\section*{Acknowledgments}

\noindent
We would like to thank Gunjan Baid, Nicole Brichtova, Sara Mahdavi, Kathy Meier-Hellstern, Zarana Parekh, Anusha Ramesh, Tris Warkentin, Austin Waters, Vijay Vasudevan for their generous help through the course of the project. We give thanks to Igor Karpov, Isabel Kraus-Liang, Raghava Ram Pamidigantam, Mahesh Maddinala, and all the anonymous human annotators for assisting us to coordinate and complete the human evaluation tasks. We are grateful to Huiwen Chang, Austin Tarango, Douglas Eck for reviewing the paper and providing feedback. Thanks to Erica Moreira and Victor Gomes for help with resource coordination. Finally, we would like to give our thanks and appreciation to the authors of DALL-E 2~\cite{ramesh-dalle2} for their permission for us to use the outputs from their model for research purposes.

{\small
\bibliographystyle{ieee_fullname}
\bibliography{main}
}

\clearpage
\appendix

\section{Appendix}
\label{sec:appendix}

\subsection{Human Evaluation Results}

\begin{figure}[t]
    \includegraphics[width=\linewidth]{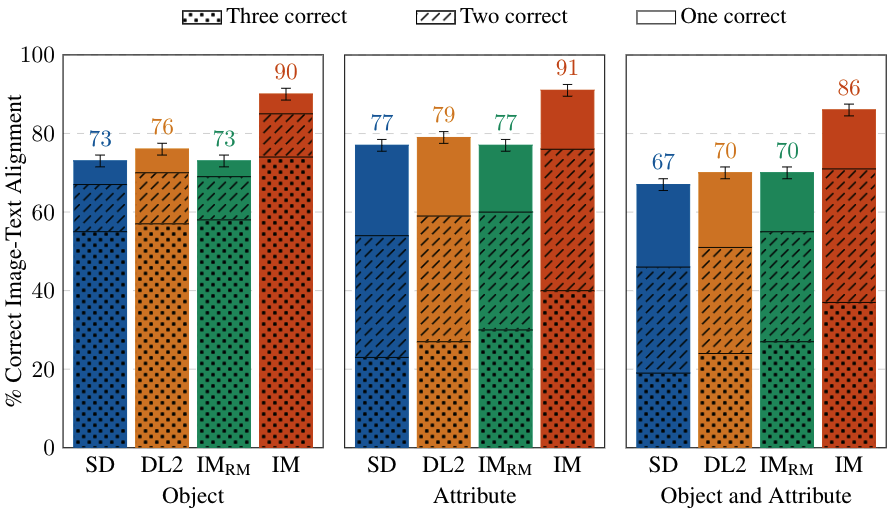}
    \vspace{-7mm}
    \caption{Single-image human evaluations on EditBench Mask-Rich illustrating the \textit{number of objects and attributes} correct. Comparing \im{} vs. \imrm{}, object-masking improves the rendering of objects and attributes as well as attribute binding.  
    }
    \label{fig:mask-rich-breakdown}
\end{figure}

\noindent
In this section we include further analysis and additional details of the human evaluation results reported in the main paper. In Fig.~\ref{fig:mask-rich-breakdown} we provide a breakdown of single-image human evaluations for the \eb{} Mask-Rich prompts, illustrating the average proportion of objects and attributes correctly rendered for each model. Focusing on a conservative bar for performance, correctly rendering \textit{at least one} of the three objects-attribute pairs in a Mask Rich prompt, \imagenator{} achieves over 85\% across comparisons. This further supports the conclusions of the main paper, object-masking (\im{} vs. \imrm{}) improves object rendering, attribute rendering, and attribute binding (i.e., object and attribute both correct, far right column).

\begin{figure}[ht]
    \centering
    \includegraphics[width=\linewidth]{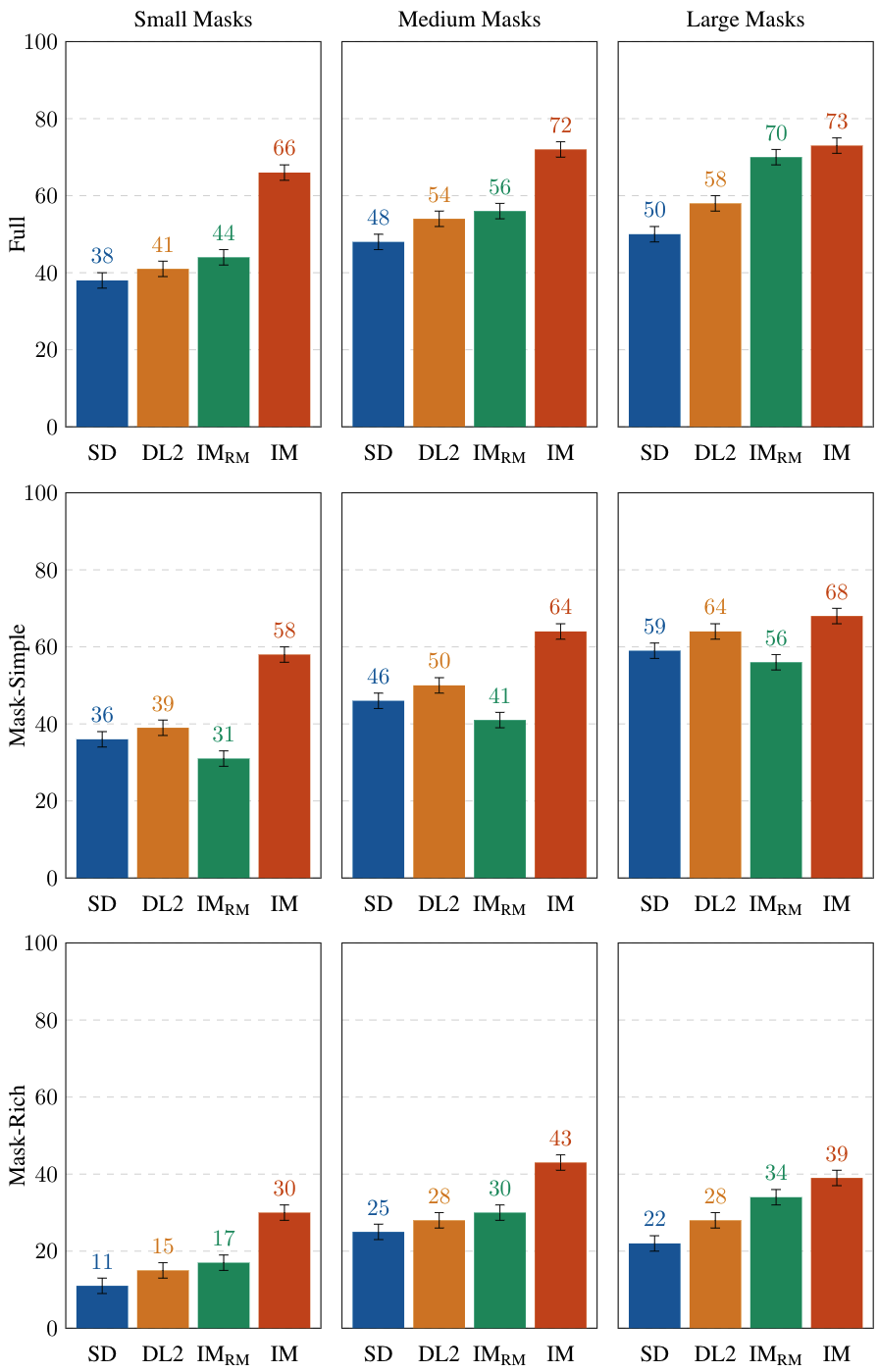}
    \vspace{-7mm}
    \caption{Single-image human evaluations on EditBench by \textit{mask size} (columns) and \textit{prompt type} (rows).
    \imagenator{} is preferred in all comparisons and object-masking during training is particularly beneficial for small masks (\im{} vs. \imrm{}).}
    \label{fig:heval-mask-area-ratio}
\end{figure}

\paragraph{Impact of Mask Size.} In Fig.~\ref{fig:heval-mask-area-ratio} we report single-image human evaluations by mask size (Small, Medium and Large). In general, performance trends are consistent with the aggregate results, however object-masking (\imrm{} vs. \im{}) is more beneficial for small and medium masks than with large masks. For reference, Fig.~\ref{fig:mask-area-ratio-example} provides examples of the different mask sizes in each bucket. 

\paragraph{Single image vs. Side-by-Side.} A key consideration in evaluating text alignment was whether to compare model outputs \emph{side-by-side}, in keeping with prior works \cite{nichol-glide,blendeddiffusion,kim2021diffusionclip,zhou-yufan-22}, or whether to evaluate each model separately -- judging a \emph{single image} at a time. We report both but focus primarily on single-image evaluations for two reasons:
\begin{itemize}
    \item \textit{Fine-grained Evaluation}. While it is reasonable to ask the annotators simple comparative questions such as \emph{which image matches the caption better?}, the task becomes more cognitive taxing and prone to error when multiple attributes and objects are compared \cite{dolnicar-11}.
    \item \textit{Avoiding Combinatorial Explosion}. The single image format facilitates pairwise comparison without eliciting judgments for a (often impractically) large number of model pairs when more models are investigated. In addition, we avoid exposing annotators to the same outputs multiple times, which might introduce exposure bias.
\end{itemize}
While the single image evaluation format may be subject to calibration biases, i.e., not all annotators will have the same threshold for judging correctness, we control for this by ensuring that all model evaluations are performed in a batch presented in random order to a large pool of annotators. 

\paragraph{Significance of single-model human evaluation.} We include 95\% confidence error bars calculated with bootstrap resampling in all single-model human evaluations (Figs. 7, 9 and 10). If error bars do not overlap, the difference in scores is significant. In particular, the difference between Imagen Editor and the other models is significant in all cases. We confirmed this, as suggested, using the one-sided, two-sample proportions z test. Illustratively, when comparing Imagen Editor's overall image-text alignment performance (Fig. 7) vs the next best model the p-values were as follows for each prompt type: Full, $P=3.5 \times {10^{-10}}$, Mask-Simple, $P=2.0 \times {10^{-8}}$, Rich, $P=5.2\times{10^{-7}}$. In Tab. 1 and Tab. 2 (correlation between automatic metrics and human evaluations) the 95\% confidence intervals calculated with bootstrap resampling are $<$1\%.

\begin{figure}[!ht]
    \centering
    \includegraphics[width=\linewidth]{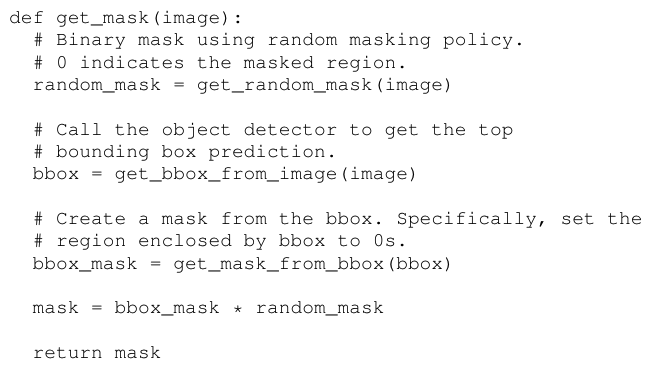}
  \caption{Bounding box based mask generation for \imagenator{}. We adapt the random mask policy used in \cite{yu2019free, sahariac-palette}.}
  \label{fig:masking_algo}
\end{figure}

\begin{figure*}
    \includegraphics[width=\linewidth]{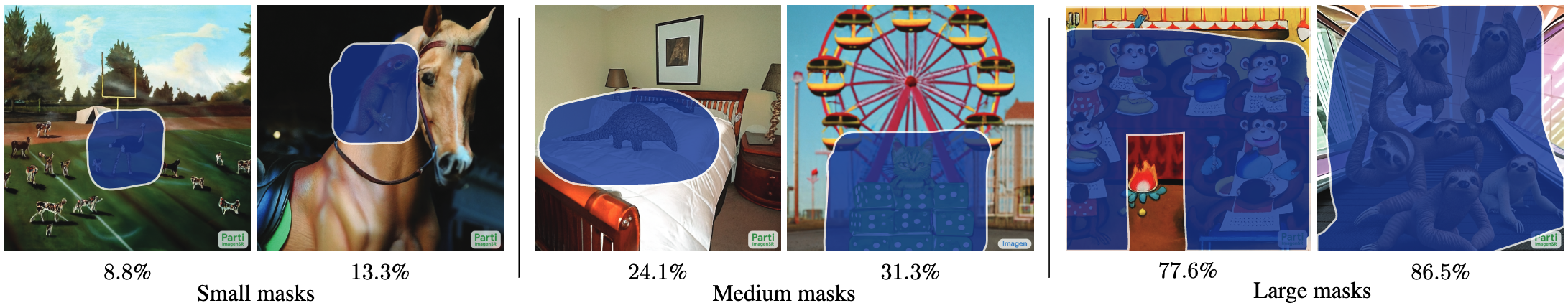}
    \caption{Examples of different mask sizes from the Small, Medium and Large buckets reported in Fig~\ref{fig:heval-mask-area-ratio}. Mask sizes were determined by binning mask-to-image area ratios into 3 quantiles as follows: Small (5.7--21.5\%), Medium (21.5--36.9\%), and Large ($>$36.9\%).}
    \label{fig:mask-area-ratio-example}
\end{figure*}

\paragraph{Sampling Strategy and Number of Evaluations.}  In \emph{single image} evaluations we evaluated 4 edited image samples for each prompt from each of the four models.  In total this gave: prompts (240) $\times$ prompt types (3) $\times$ image samples (4) $\times$ models (4) = 11,520 outputs rated by annotators.  In the \emph{side-by-side} evaluation of Mask-Rich prompts, we evaluated 3 model pairs (\imagenator{} vs. \stablediffusion{}, \dalle{} and \imagenatorrm{}), resulting in: 3 $\times$ 240 (images) $\times$ 1 (prompt types) $\times$ 3 (votes from different annotators) = 2,160 ratings. In side-by-side evaluations, an image was selected at random from the 4 samples from each model.

\paragraph{Annotators.} 
We use a total of 18 US-based annotators and the evaluation load was spread approximately equally. Each annotator spent roughly $\sim$30s per prompt.

\begin{figure*}
    \centering
    \begin{subfigure}[b]{\textwidth}
    \centering
    \includegraphics[width=0.8\linewidth]{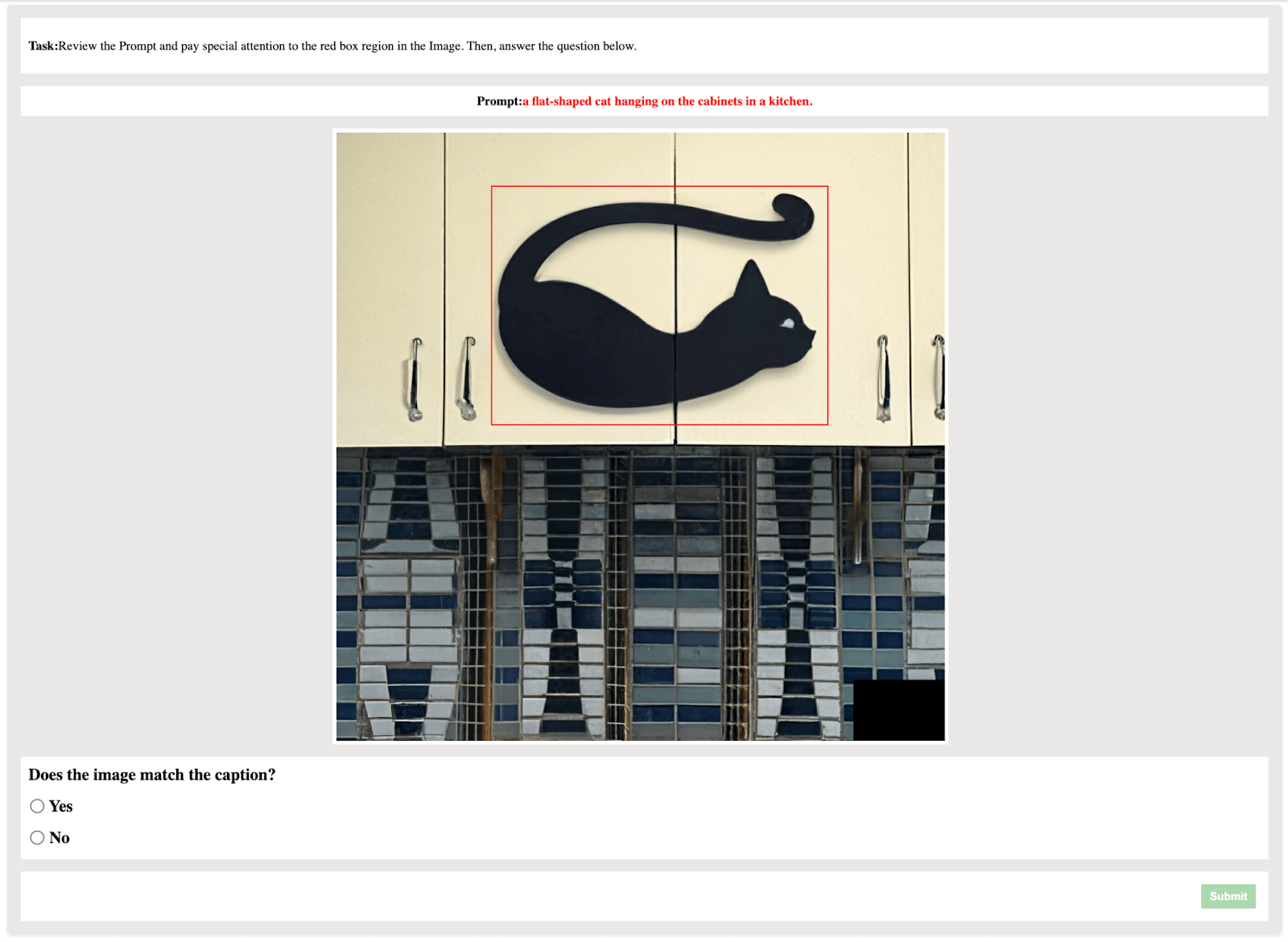}
    \caption{\full prompt single image evaluation with binary selection for overall text-image alignment.}
    \label{fig:human-ui-full}
    \end{subfigure}

    \begin{subfigure}[b]{\textwidth}
    \centering
    \includegraphics[width=0.8\linewidth]{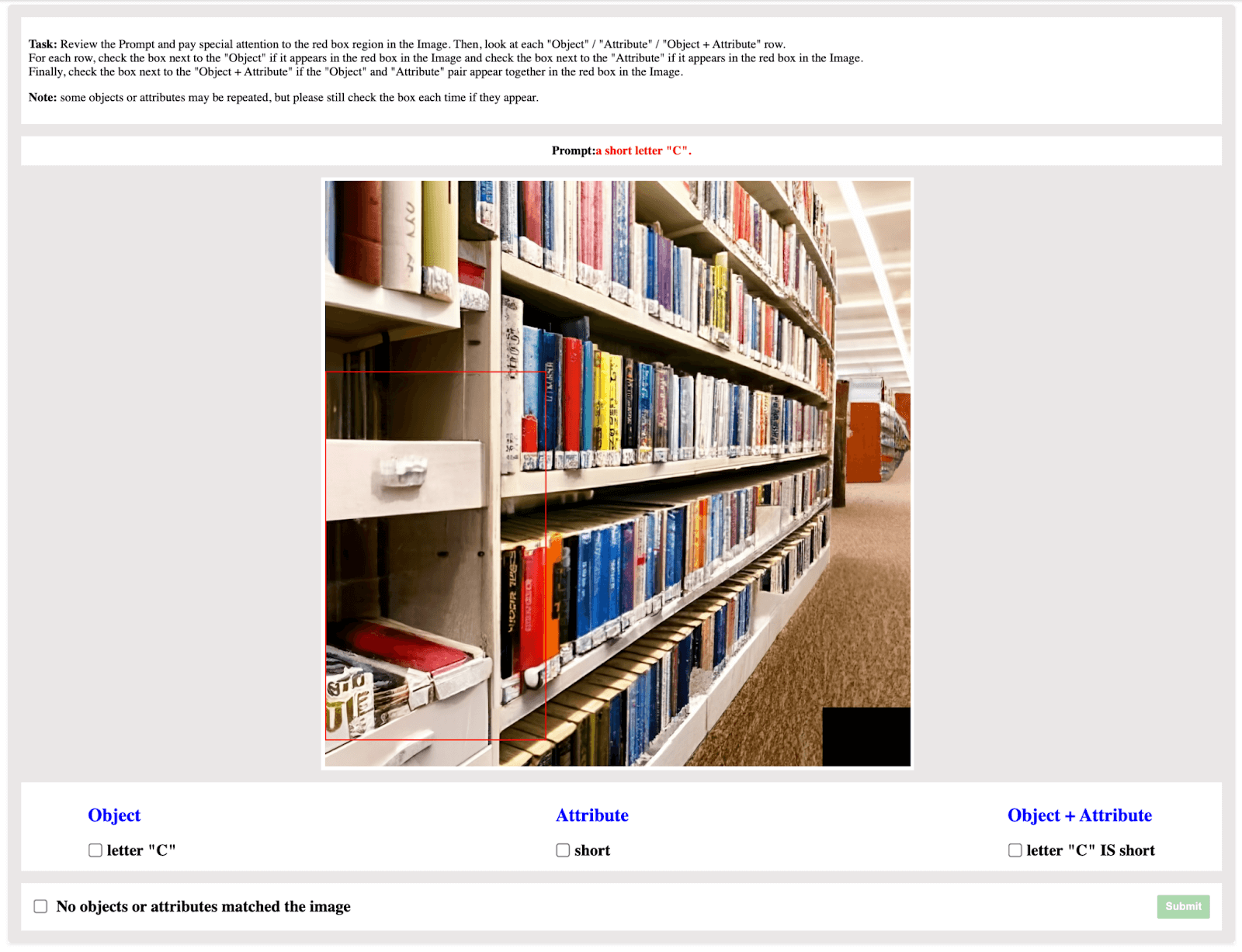}
    \caption{\simple prompt evaluation where annotators assess existence of the correct object and attribute separately and together to measure correct attribute binding.}
    \label{fig:human-ui-simple}
    \end{subfigure}
    
    \caption{Crowdsourcing UI illustration: \full prompts \& \simple prompts.}
    \label{fig:human-ui-full-and-simple}

\end{figure*}

\begin{figure*}
    \centering
    
    \begin{subfigure}[b]{0.8\textwidth}
    \includegraphics[width=\linewidth]{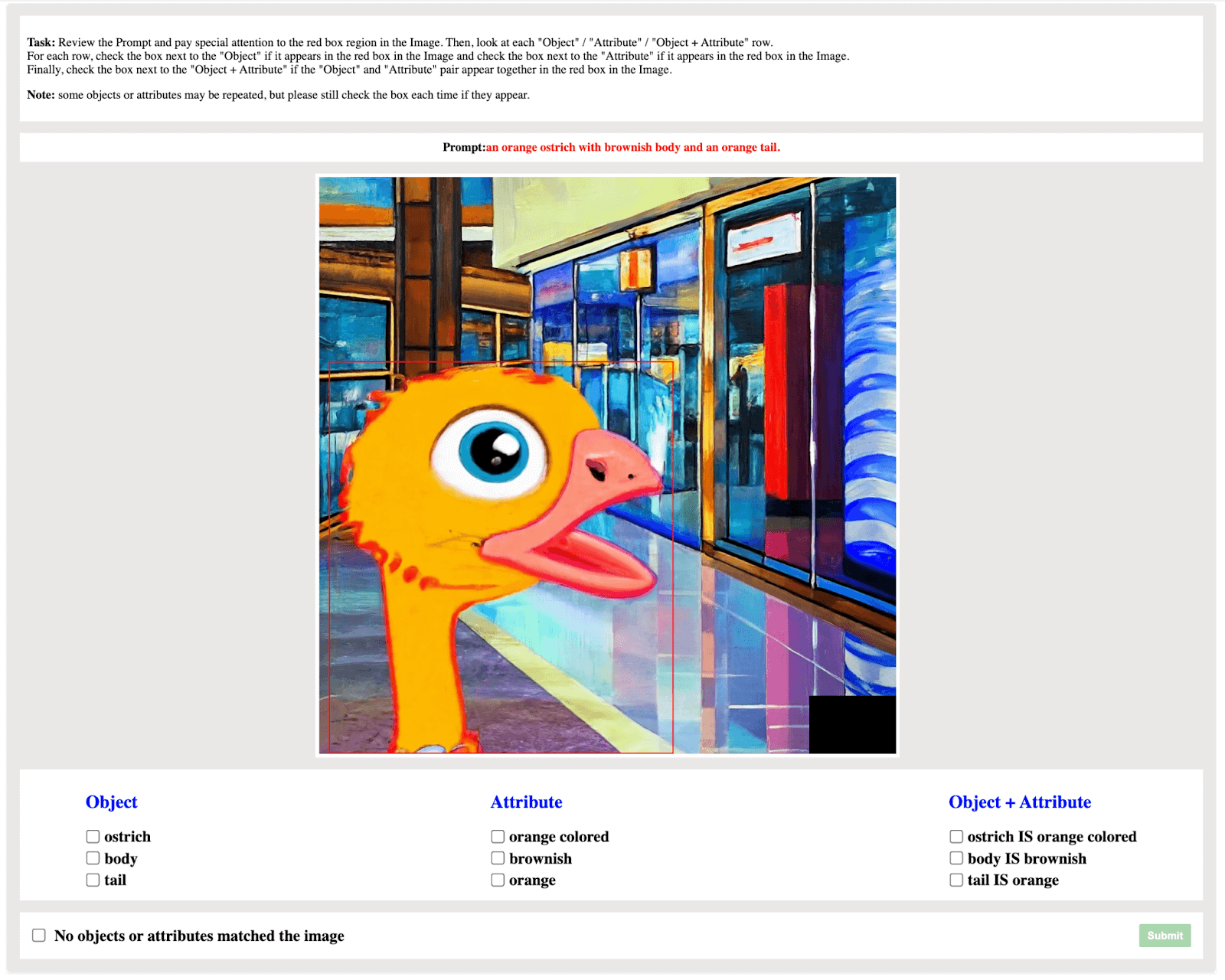}
    \caption{\rich prompt evaluation, which is similar to \simple (Fig. \ref{fig:human-ui-simple}) but involves 3 pairs of attributes and objects.}
    \label{fig:human-ui-rich}
    \end{subfigure} 
    
    \begin{subfigure}[b]{0.8\textwidth}
    \includegraphics[width=\linewidth]{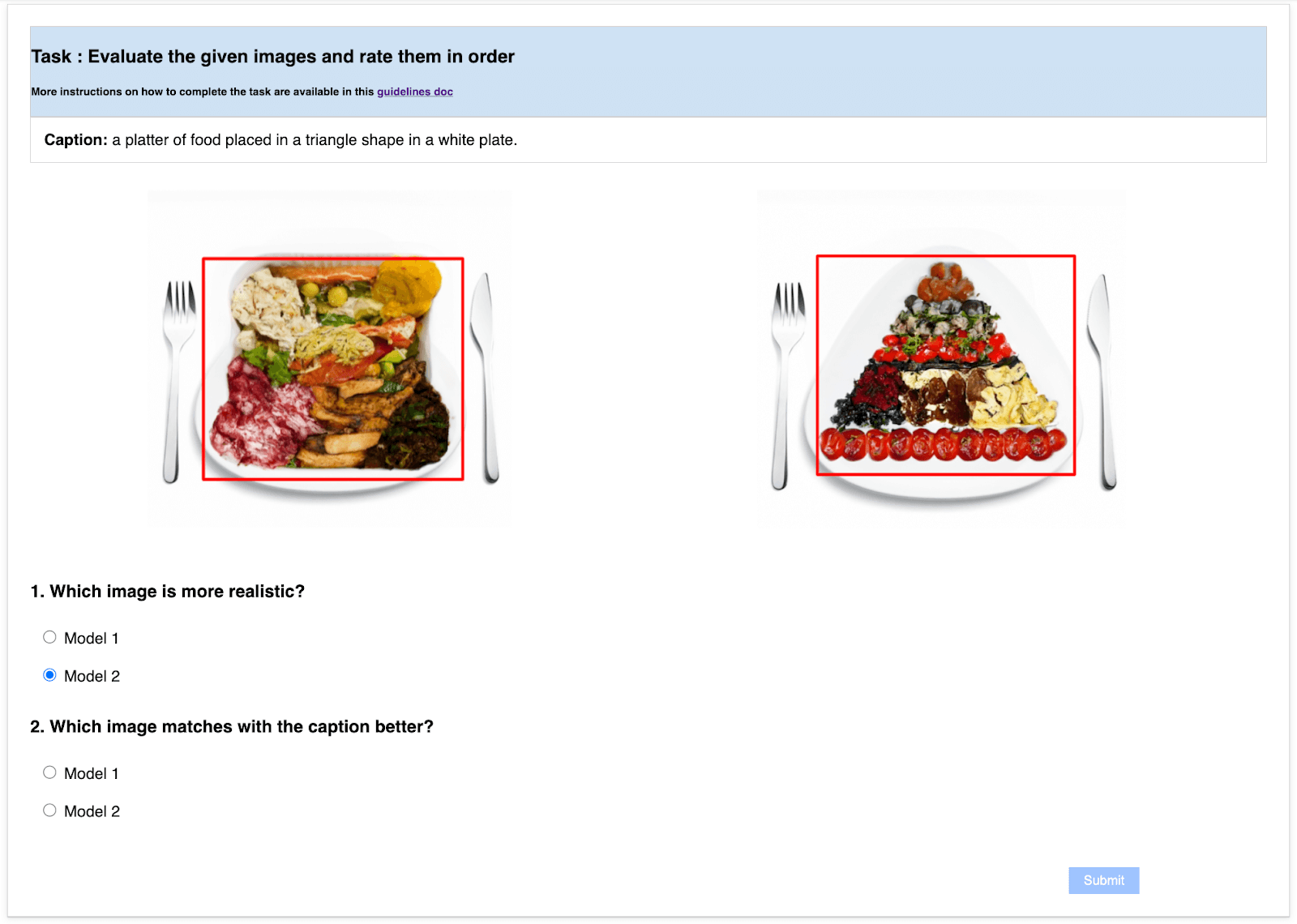}
    \caption{Side-by-Side evaluation. The second (text-image alignment) question only appears after the first (realism) question is answered.}
    \label{fig:human-ui-sxs}
    \end{subfigure} 
    
    \caption{Crowdsourcing UI illustration: \rich prompts \& \emph{Side-by-side} evaluations.}
    \label{fig:human-ui-rich-and-sxs}
    
\end{figure*}

\paragraph{Crowdsourcing UI.} 
We illustrate the actual interface used in our human evaluations in Figs. \ref{fig:human-ui-full-and-simple}, \ref{fig:human-ui-rich-and-sxs}. 

\subsection{Imagen Editor object masking}

\noindent
We apply a bounding box based masking strategy for \imagenator, which is an adaptation of random mask policy used in previous work. During training the mask is the union of a random mask and an object detection bounding box as described in Fig. \ref{fig:masking_algo}.

\subsection{Examples and Failure Cases}

\begin{figure*}
\centering
\includegraphics[width=\linewidth]{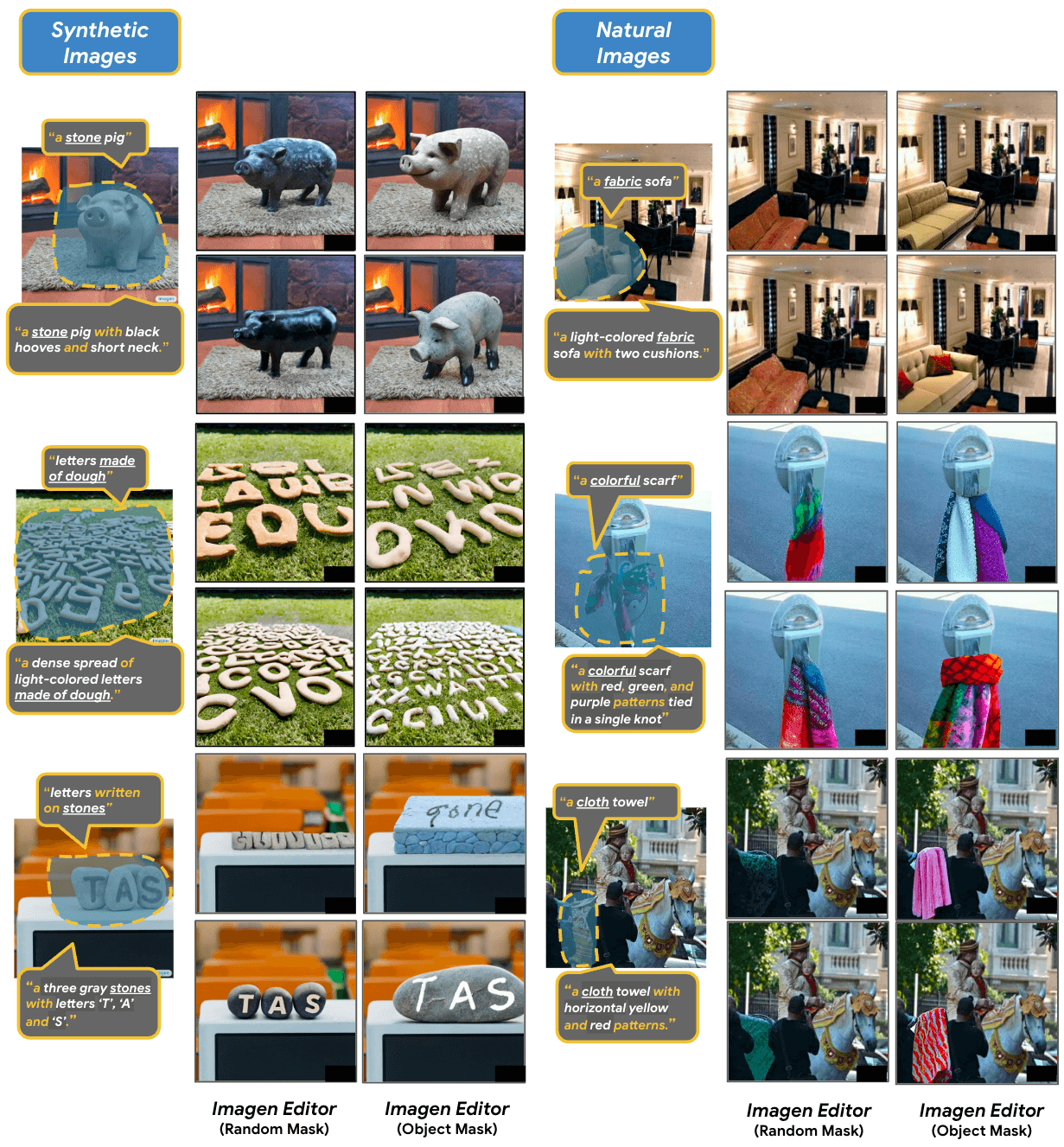}
\caption{Additional examples comparing the random and object-masking strategies on \simple and \rich prompts. \imagenator{} is substantially more robust at handling richer attribute/object specifications, as confirmed by human evaluations.
}
\label{fig:example-splash-im-v1-vs-v2}
\end{figure*}

\begin{figure*}
\centering
\includegraphics[width=\linewidth]{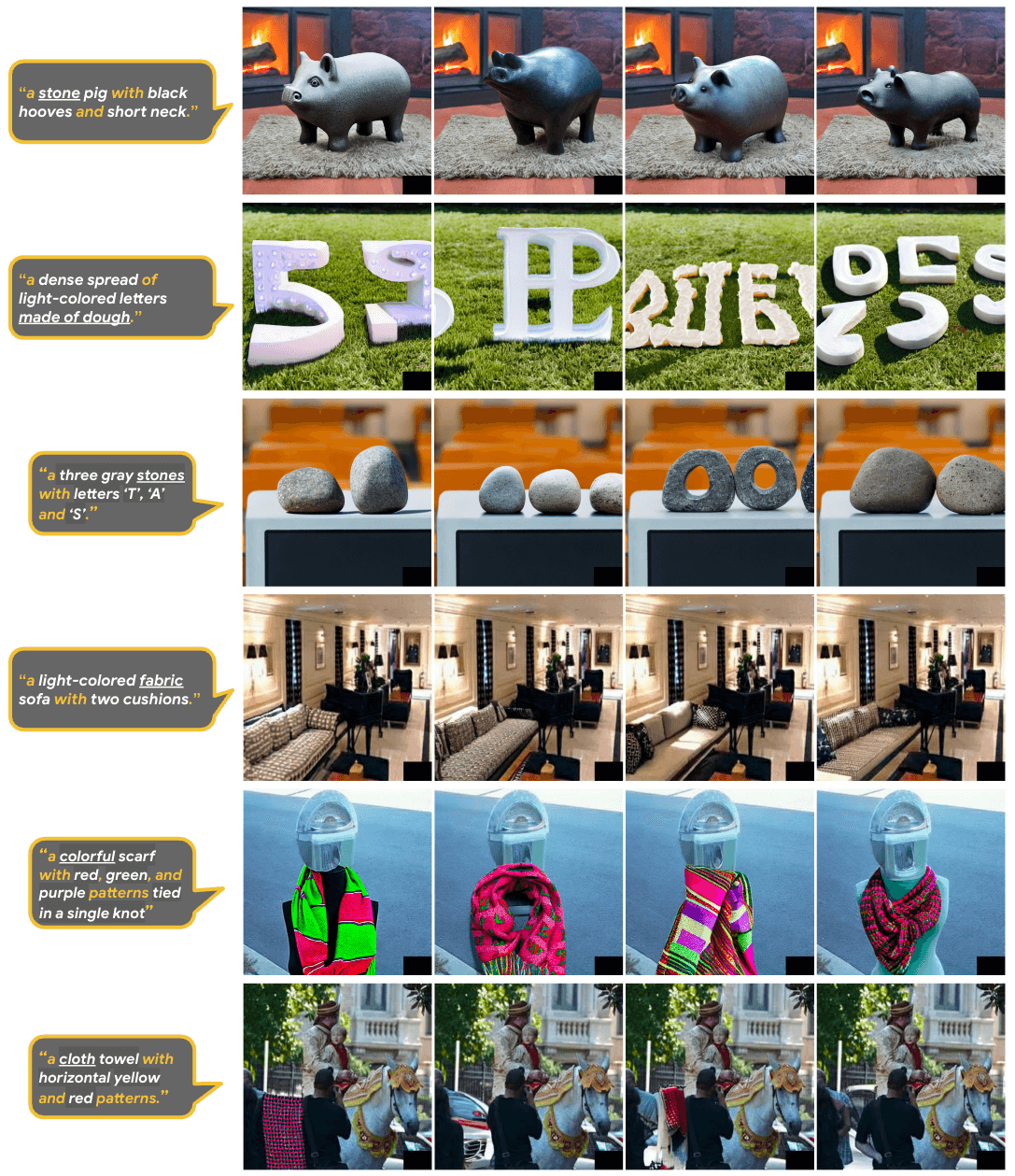}
\caption{Stable Diffusion examples.}
\label{fig:appendix-example-sd}
\end{figure*}

\begin{figure*}
\centering
\includegraphics[width=\linewidth]{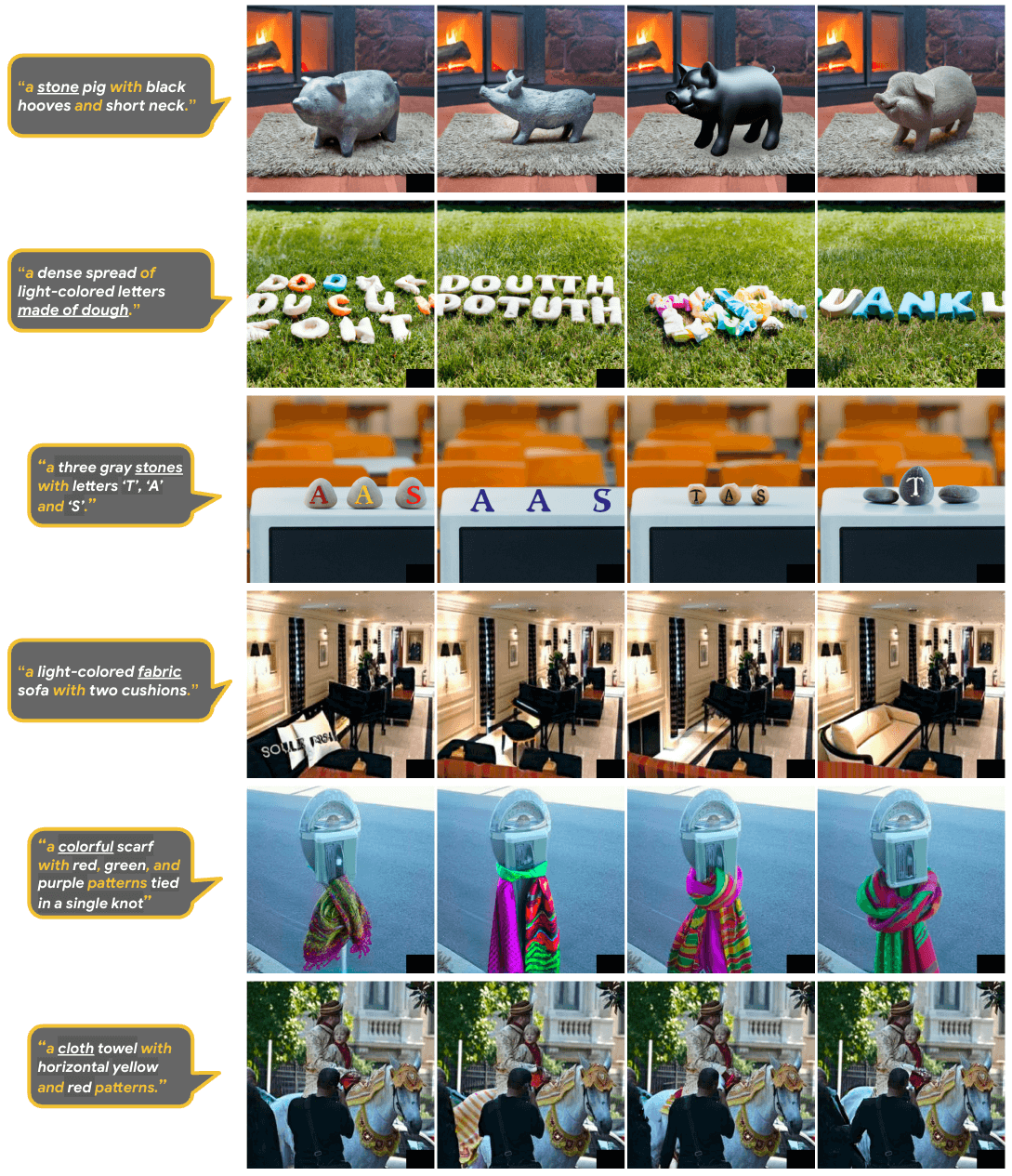}
\caption{DALL-E 2 examples.}
\label{fig:appendix-example-dl}
\end{figure*}

\begin{figure*}
\centering
\includegraphics[width=\linewidth]{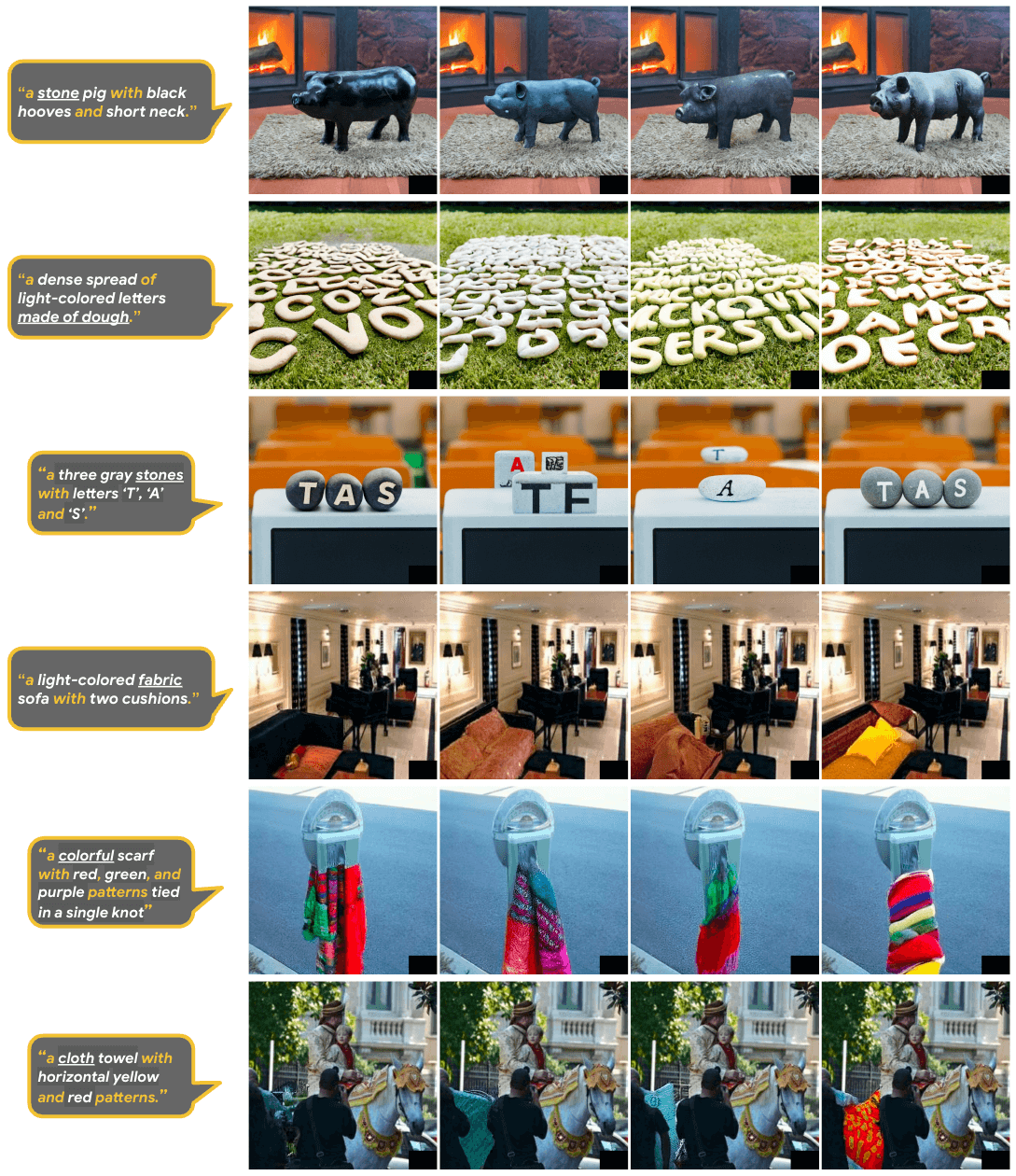}
\caption{\imagenator{} (random masking) examples.}
\label{fig:appendix-example-imrm}
\end{figure*}

\begin{figure*}
\centering
\includegraphics[width=\linewidth]{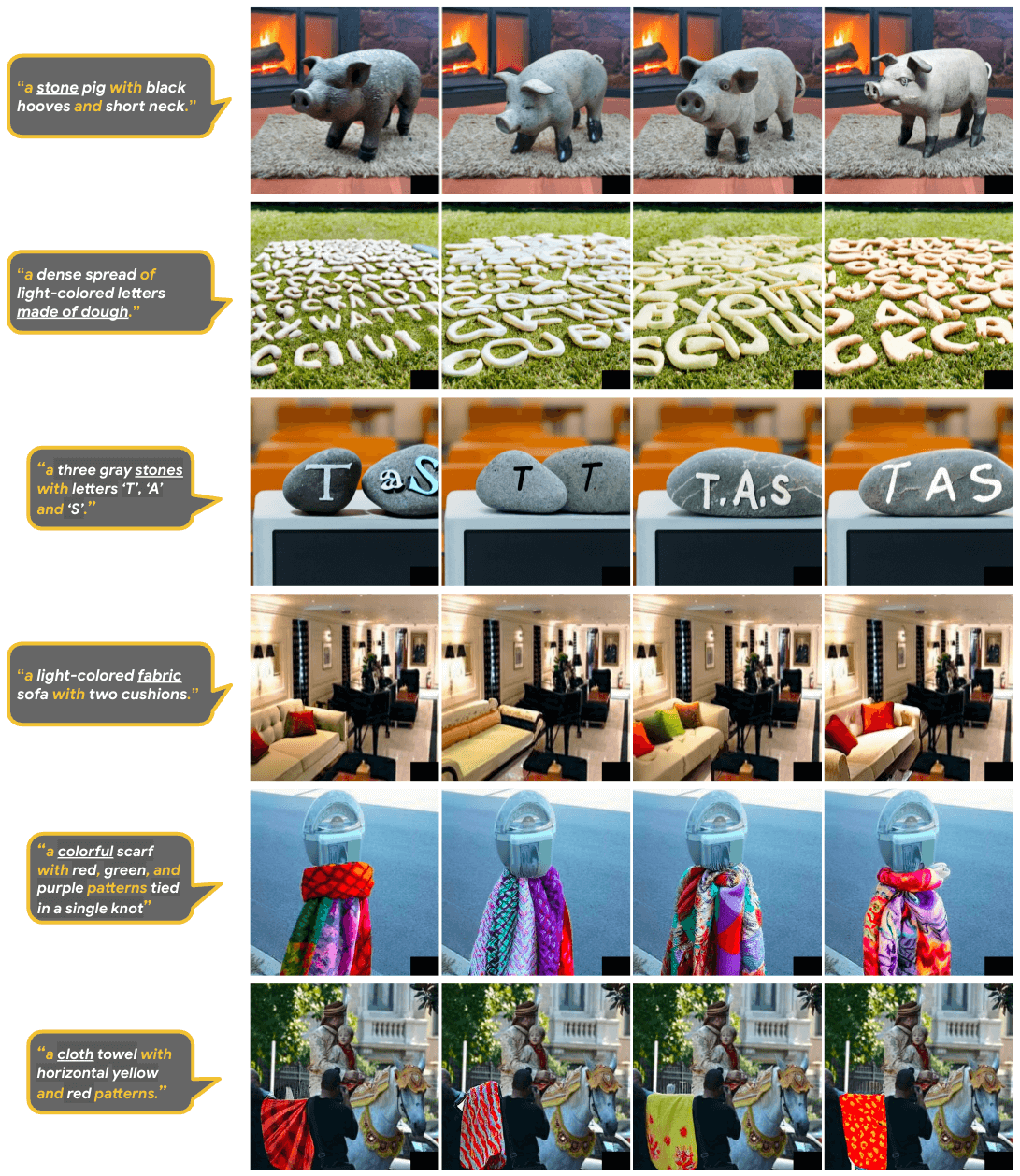}
\caption{\imagenator{} (object masking) examples.}
\label{fig:appendix-example-im}
\end{figure*}

\noindent
In Fig.~\ref{fig:example-splash-im-v1-vs-v2} we provide further examples comparing outputs from \imagenator when trained with object-masking vs. random masking. We find that object masking makes the model noticeably more robust when handling richer prompts with more details of objects and their attributes. To illustrate the variety of samples evaluated from each model, in Figs.~\ref{fig:appendix-example-sd}--\ref{fig:appendix-example-im} we illustrate sampled outputs from \stablediffusion{}, \dalle{}, \imagenatorrm{} and \imagenator{} respectively. 

\begin{figure*}
\centering
\includegraphics[width=\linewidth]{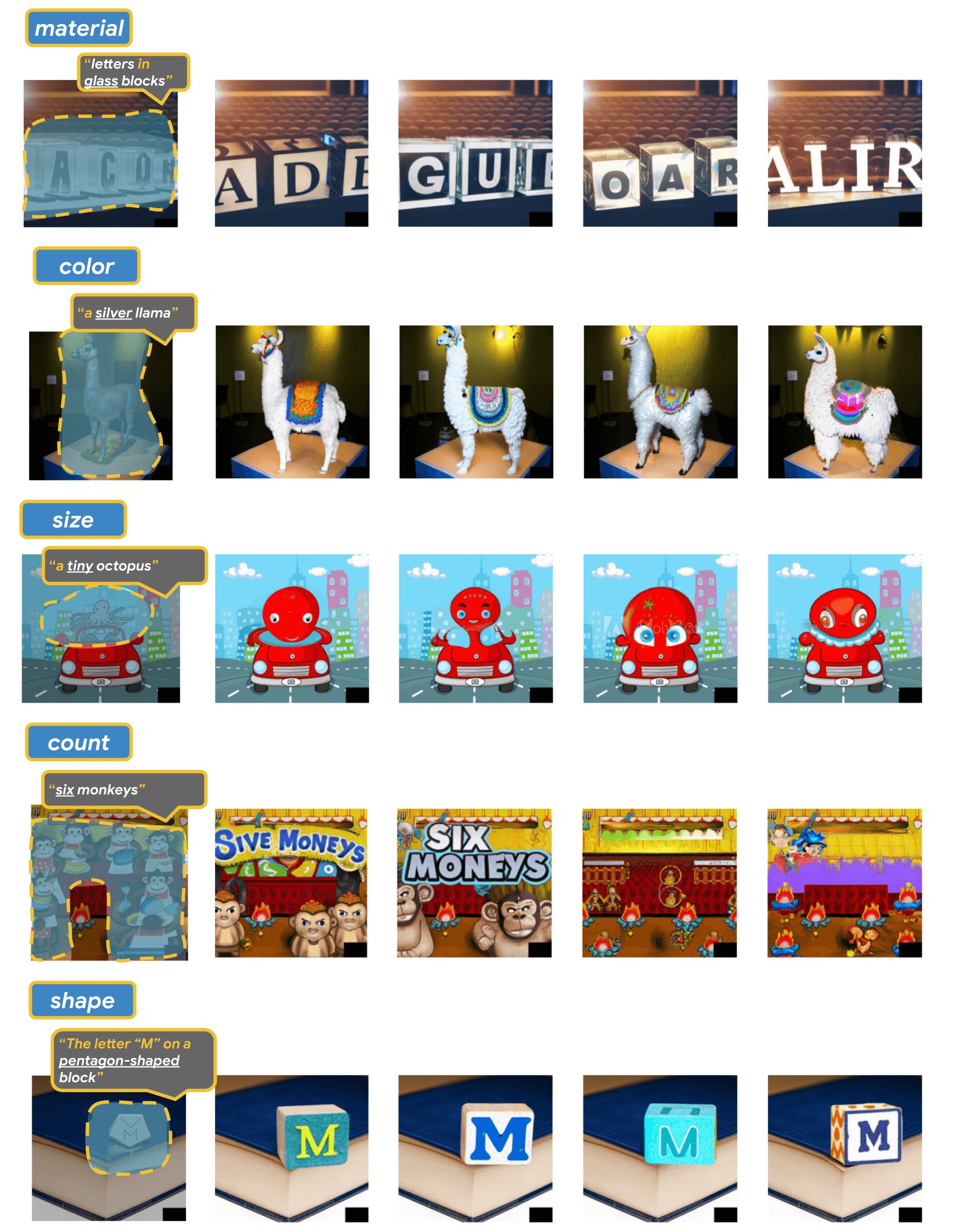}
\caption{\imagenator{} failure cases by attribute. \textbf{Material} - the blocks don't quite have the transparency property you'd expect of letters encased in a glass box. \textbf{Color} - the llama is not quite silver colored in any case. The object ``llama'' is object type = ``uncommon'' denoting that a ``silver'' llama is likely out of distribution for the model and therefore, more challenging.  \textbf{Size} - the octopus is never quite ``tiny'' this is wrong in at least two ways: not absolutely (with respect to the image size), nor relatively (compared to the car). \textbf{Count} - count is among the more challenging attributes and anecdotally, models are often off by 50\% or more.  \textbf{Shape} - all image samples revert to the standard shape for a block: a cube.} 
\label{fig:appendix-error-analysis}
\end{figure*}

\paragraph{\imagenator{} Failure Cases.} In Fig. \ref{fig:appendix-error-analysis}, we further explore \imagenator{} failure cases.  We focus on attribute types as Fig. \ref{fig:mask-rich-breakdown} shows that, even in the case of more complex, Mask Rich prompts, models are relatively strong at getting the majority of objects mentioned correct.  As is consistent with our breakdown of Mask-Simple prompts by Attribute type,
a qualitative review of \imagenator{} failure cases on Mask-Simple prompts supports that \imagenator{} is fairly strong on \textit{color} and \textit{material}.  Where there were failure cases the objects or the colors tended to be uncommon (i.e. ``butter-colored'' letters or ``silver'' llama in the figure).  \textit{Size} and \textit{shape} are admittedly often more challenging because they can be more ambiguous.  Yet still there are a handful of cases where the size attribute appears to be ignored (i.e. ``tiny octopus'' in figure).  In almost all cases, some object and often it's more common attributes are inpainted (an example of this is the ``pentagon-shaped block'' instead of a cube-shaped block).  Finally, \textit{count} is notoriously challenging and the most clear failure case of the various models.  Rarely do they render too many objects. Almost always they render far too few, often over 50\% of objects are missing.

\end{document}